%% file: main.tex
\documentclass[10pt,journal,compsoc]{IEEEtran}
\ifCLASSOPTIONcompsoc
  \usepackage[nocompress]{cite}
\else
  \usepackage{cite}
\fi

\usepackage{algorithm}
\usepackage{algorithmic}
\usepackage{amssymb}
\usepackage{multirow}
\usepackage{pifont}
\usepackage{makecell}
\usepackage{amsmath}
\usepackage{color}
\usepackage{xcolor}
\usepackage{arydshln}
\usepackage{newfloat}
\usepackage{listings}
\usepackage{times}
\usepackage{latexsym}
\usepackage{graphicx}
\usepackage{array}
\usepackage{booktabs}
\usepackage{enumitem}
\usepackage{ragged2e}
\newcommand{\zqh}[1]{{\color{black} #1}}

\begin{document}

\title{Knowledge Graph Augmented Network Towards Multiview
 Representation Learning for Aspect-based Sentiment Analysis}

\author{Qihuang~Zhong,~\IEEEmembership{Member,~IEEE,}
        Liang~Ding,~\IEEEmembership{Member,~IEEE,}
        Juhua~Liu,~\IEEEmembership{Member,~IEEE,}
        Bo~Du,~\IEEEmembership{Senior~Member,~IEEE,}
        Hua Jin,
        and~Dacheng~Tao,~\IEEEmembership{Fellow,~IEEE}
\IEEEcompsocitemizethanks{\IEEEcompsocthanksitem Q. Zhong and B. Du are  with the National Engineering Research Center for Multimedia Software, Institute of Artificial Intelligence, School of Computer Science and Hubei Key Laboratory of Multimedia and Network Communication Engineering, Wuhan University, Wuhan, China (e-mail: zhongqihuang@whu.edu.cn; dubo@whu.edu.cn).

\IEEEcompsocthanksitem J. Liu is with the Research Center for Graphic Communication, Printing and Packaging, Institute of Artificial Intelligence, Wuhan University, Wuhan, China (e-mail: liujuhua@whu.edu.cn).

\IEEEcompsocthanksitem L. Ding and D. Tao are with the JD Explore Academy at JD.com, Beijing, China (e-mail: dingliang1@jd.com; dacheng.tao@jd.com).

\IEEEcompsocthanksitem H. Jin is with the affiliated hospital of Kunming University of Science and Technology, Kunming, China (e-mail: jinhuakm@163.com).

\IEEEcompsocthanksitem Corresponding Authors: Juhua~Liu (e-mail: liujuhua@whu.edu.cn), Bo Du (e-mail: dubo@whu.edu.cn)
}
\thanks{This work was done during Qihuang Zhong’s internship at JD Explore Academy. 
Our source code and final models are publicly available at https://github.com/WHU-ZQH/KGAN}.}

\markboth{Journal of \LaTeX\ Class Files, December~2021}%
{Shell \MakeLowercase{\textit{et al.}}: Bare Demo of IEEEtran.cls for Computer Society Journals}

\IEEEtitleabstractindextext{%
\begin{abstract}
\justifying{Aspect-based sentiment analysis (ABSA) is a fine-grained task of sentiment analysis. To better comprehend long complicated sentences and obtain accurate aspect-specific information, linguistic and commonsense knowledge are generally required in this task.
However, most current methods employ complicated and inefficient approaches to incorporate external knowledge, \textit{e.g.}, directly searching the graph nodes. Additionally, the complementarity between external knowledge and linguistic information has not been thoroughly studied. To this end, we propose a knowledge graph augmented network (\textsc{KGAN}), which aims to effectively incorporate external knowledge with explicitly syntactic and contextual information. In particular, KGAN captures the sentiment feature representations from multiple different perspectives, \textit{i.e.}, context-, syntax- and knowledge-based. First, KGAN learns the contextual and syntactic representations in parallel to fully extract the semantic features. Then, KGAN integrates the knowledge graphs into the embedding space, based on which the aspect-specific knowledge representations are further obtained via an attention mechanism. Last, we propose a hierarchical fusion module to complement these multi-view representations in a \textit{local-to-global} manner. Extensive experiments on \zqh{five} popular ABSA benchmarks demonstrate the effectiveness and robustness of our KGAN. Notably, with the help of the pretrained model of RoBERTa, KGAN achieves a new record of state-of-the-art performance among all datasets.
} 
\end{abstract}

}

\maketitle

\IEEEdisplaynontitleabstractindextext

\IEEEpeerreviewmaketitle

\IEEEraisesectionheading{\section{Introduction}
\label{sec:introduction}}

\begin{figure}[]
	\centering
	\includegraphics[width=0.47\textwidth]{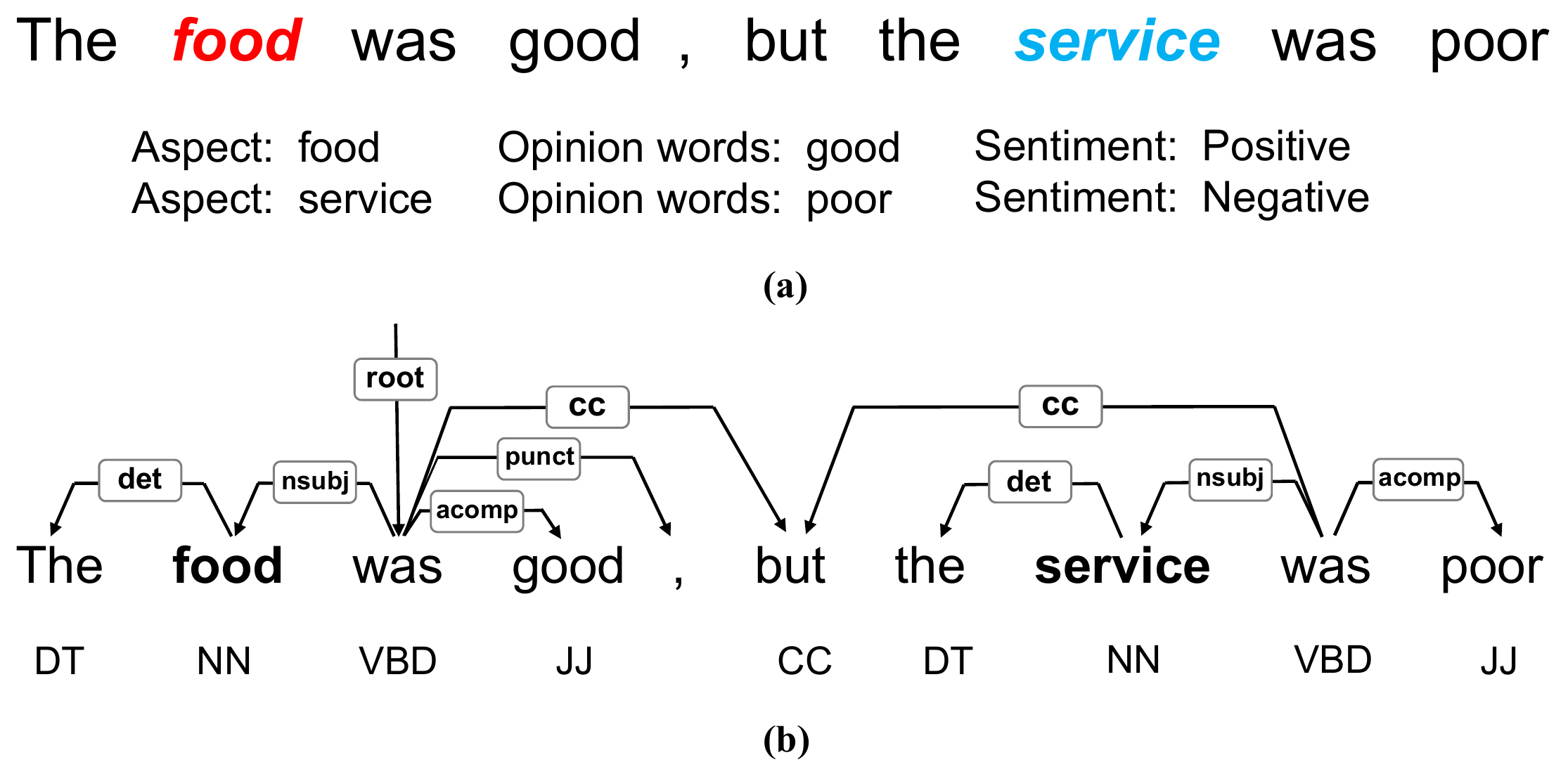} 
	\caption{(a) An example sentence of the ABSA task from the restaurant reviews. There are two aspects with opposite sentiment polarities in this sentence. (b) Illustration of the dependency parsing result.}
	\label{Fig00}
\end{figure}

\IEEEPARstart{A}{s} a fine-grained task of sentiment analysis, aspect-based sentiment analysis (ABSA) has grown to be an active research task in the community of natural language understanding (NLU) \cite{liu2012survey,zhang2018deep,schouten-tkde2016}. In particular, ABSA refers to judging the sentiment polarities (\textit{e.g.}, positive, neutral, and negative) towards the given aspects, which are usually the target entities appearing in the sentence \cite{pontiki-etal-2014-semeval}. Taking the sentence ``The \textbf{\textit{food}} was good, but the \textbf{\textit{service}} was poor." as an example, as shown in Fig.~\ref{Fig00}(a), the goal of ABSA is to predict the polarities ``positive" and ``negative" for the aspects \textbf{\textit{food}} and \textbf{\textit{service}}, respectively.

Recent ABSA modeling approaches are mainly based on deep neural networks (DNNs) owing to the capability of automatically extracting semantic features \cite{schouten2015survey}. Specifically, based on the type of learned feature representations, existing DNNs for ABSA can be classified into two groups: context-based methods \cite{li2018transformation,xu2020target,xu-etal-2020-aspect} and syntax-based methods \cite{zhang-etal-2019-aspect,hou-etal-2021-graph,li-etal-2021-dual-graph}. Context-based methods first employ convolutional neural networks (CNNs) or long short-term memory networks (LSTMs) to extract the features of aspects and context words and then use the attention mechanism to capture the aspect-specific contextual representations. In addition to context-based methods, syntax-based methods attempt to model the nonlocal dependency trees (a case in point is shown in Fig.~\ref{Fig00}(b)) of sentences with graph neural networks, \textit{e.g.}, graph convolutional networks (GCNs) to encode the syntactic information and syntactically connect the aspects with related opinion words \cite{wang2020relational}.

More recently, given effective knowledge, \textit{e.g.}, linguistic and commonsense, for representation approaches in NLU tasks~\cite{yasunaga2021qa, liu2018entity,Hu-tkde-2018}, researchers employ external knowledge to augment the semantic features in ABSA models~\cite{ma2018targeted,zhou2020sk,zhao2021knowledge,zhang-tkde-2021}. However, they make extensive modifications to model structures or objectives to encode the different kinds of knowledge, limiting the applicability of their methods to a broader range of tasks and knowledge types. For example, \zqh{Zhou \textit{et al.}~\cite{zhou2020sk}} directly utilized the words (\textit{w.r.t. aspect terms in sentences}) in knowledge graphs as the seed nodes and selected the related nodes to construct the subgraphs. \zqh{While these subgraph-based methods~\cite{zhou2020sk, islam2022ar} have achieved remarkable performance, there are still some problems, \textit{e.g.}, the process of constructing subgraphs is usually relatively complex and would bring more computation, especially when there are many aspect terms. Hence, we attempt to integrate external knowledge from a different perspective.}

In this paper, we propose a novel knowledge graph augmented network, namely, KGAN, to integrate external knowledge for boosting the performance of ABSA task. In general, KGAN employs three parallel branches to learn the feature representations from multiple perspectives (\textit{i.e.}, context-, syntax- and knowledge-based). The contextual and syntactic branches are used to extract the explicit context and syntax information from the labeled ABSA data, respectively, as most existing ABSA models do. More specifically, in the knowledge branch, unlike the above previous methods that usually employ complicated approaches to encode the knowledge, we recast them with a simpler and more efficient strategy to incorporate the external knowledge. In practice, instead of directly operating on graph-structure data, we first integrate external knowledge graphs into low-dimensional continuous embeddings, which can be simply and efficiently used to represent sentences and aspects. Then, based on the knowledge embeddings, a soft attention mechanism is utilized to capture the aspect-specific knowledge representations. As a result, we can obtain multiple representations that establish the relations between aspects and opinion words from different views. To take full advantage of the complementarity of these multiview representations, we introduce a novel hierarchical fusion module to effectively fuse them.

We conduct a comprehensive evaluation of KGAN on SemEval2014 (\textit{i.e.}, Laptop14 and Restaurant14), \zqh{SemEval2015 (\textit{i.e.}, Restaurant15), SemEval2016 (\textit{i.e.}, Restaurant16)} and Twitter benchmarks. The experimental results show that KGAN achieves comparable performance compared to the prior SOTA model with the GloVe-based setting. Moreover, we also investigate and demonstrate the effectiveness and robustness of our KGAN in BERT- and RoBERTa-based settings. \zqh{In particular, based on RoBERTa, our model achieves the SOTA performance among all datasets in terms of accuracy and macro-F1 score. More specifically, compared to the prior SOTA models, the accuracy improvements of KGAN on Twitter, Restaurant15 and Restaurant15 datasets are up to 2.49\%, 3.28\% and 2.06\%, respectively.} Finally, we also compare KGAN with the other models in terms of latency and model size and prove that KGAN can achieve a good trade-off between efficiency and performance.

The main contributions can be summarized as follows:
\begin{enumerate}
	\item We propose a novel knowledge graph augmented network (KGAN), where different types of information are encoded as multiview representations to augment the semantic features, thus boosting the performance of ABSA.
	
	\item To achieve better complementarity between multiview features, we design a novel hierarchical fusion module to effectively fuse them. 
	
	\item Experiments on several commonly used ABSA benchmarks show the effectiveness and universality of our proposed KGAN. In combination with pretrained models, \textit{i.e.}, RoBERTa, we achieve new state-of-the-art performance on these benchmarks.
\end{enumerate}

The rest of this paper is organized as follows. In Sec.~\ref{sec:related work}, we briefly review the related works. In Sec.~\ref{sec:method}, we introduce our proposed method in detail. Sec.~\ref{sec:experiment} reports and discusses our experimental results. Lastly, we conclude our study in Sec.~\ref{sec:conclusion}.

\begin{figure*}[t]
	\centering
	\includegraphics[width=0.95\textwidth]{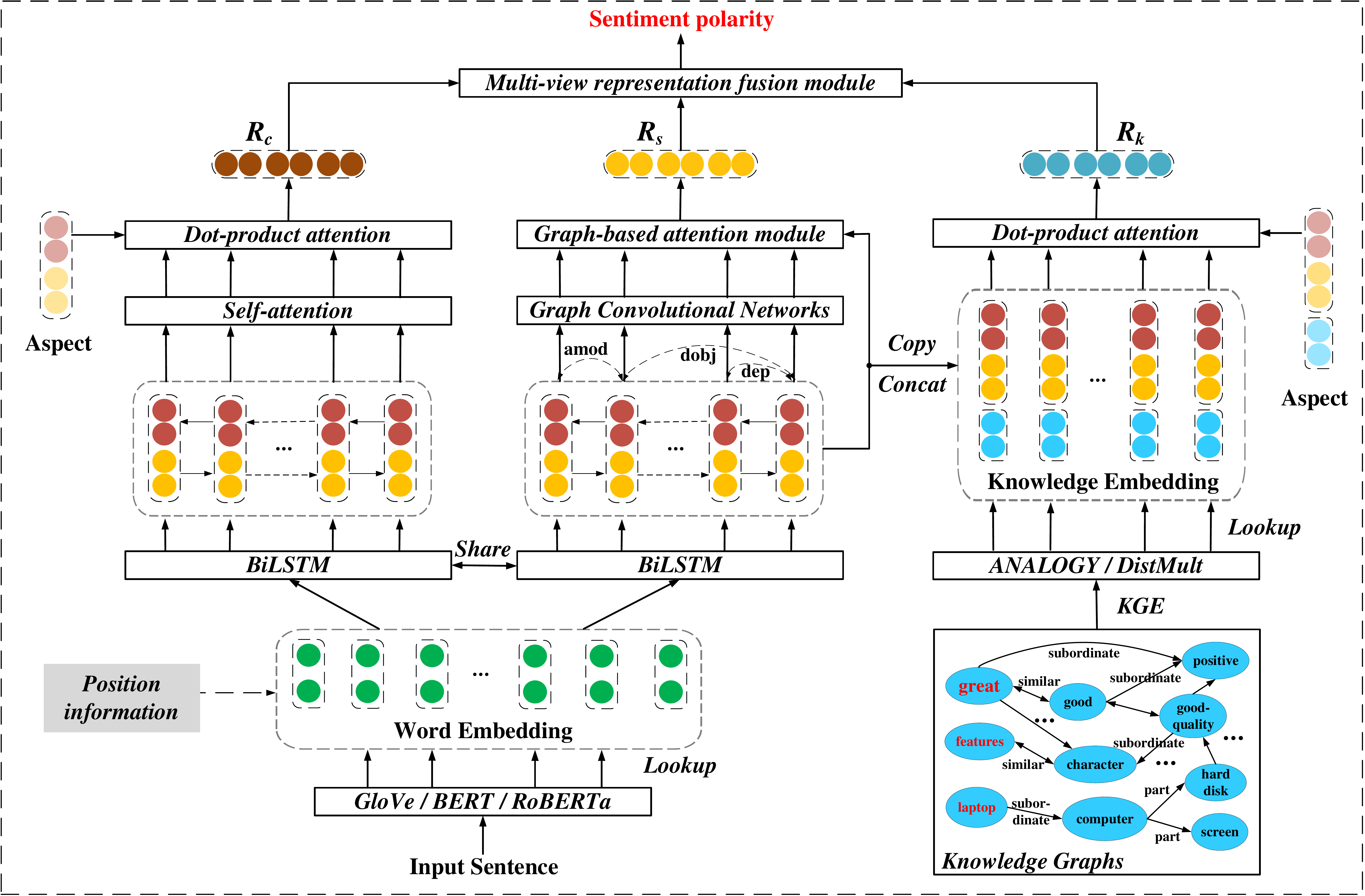} 
	\caption{The architecture of our proposed knowledge graph augmented network (KGAN), which leverages external knowledge graphs to augment contextual and syntactic information. The $R_c$, $R_s$ and $R_k$ denote the context- (left), syntax- (middle) and knowledge-based (right) representations, respectively. In the knowledge branch, ANALOGY and DistMult refer to the approaches of Knowledge Graph Embeddings (KGE). \zqh{The GloVe/BERT/RoBERTa is used to convert the sentence/aspect into word embeddings.}}
	\label{Fig0}
\end{figure*}

\section{Related Works} \label{sec:related work}
\subsection{Aspect-based Sentiment Analysis}
Benefiting from the representation learned from the training data, DNN-based ABSA models have shown promising performance compared to handcrafted feature-based models. We categorize them into two classes, \textit{e.g.}, context- and syntax-based methods.

First, considering the easily obtained contextual information, using CNNs \cite{xue2018aspect,li2018transformation,fan2018convolution,huang2019parameterized,chen2020inducing} and LSTMs \cite{tang2016effective,wang2016attention,ma2017interactive,ma2018targeted,xu2020target,jiang2019challenge} to extract the aspect-specific feature representations from context has become the mainstream approach for ABSA. In particular, owing to the ability to learn sequential patterns, the target-dependent LSTM (TD-LSTM) was proposed by Tang \textit{et al.} \cite{tang2016effective} to capture the aspect information. TD-LSTM simplifies connecting the aspect with all context words, neglecting the effect of relative opinion words. Therefore, Wang \textit{et al.} \cite{wang2016attention} improved upon the TD-LSTM by introducing an attention mechanism to explore the potential correlations between aspects and opinion words. In the study of Ma~\textit{et al.} \cite{ma2017interactive}, two separate LSTMs were used to encode the context and aspect terms, and then an interactive attention mechanism was further proposed to extract the more relevant information between the context and aspect features. 

On the other hand, considering the complexity and inefficiency of LSTM-like sequential models, many studies have attempted to employ more efficient CNNs to capture the compositional structure and n-gram features. Xue and Li \cite{xue2018aspect} proposed a gated convolution network to extract the contextual features and employed the gate mechanism to selectively output the final sentiment features. Huang and Carley \cite{huang2019parameterized} introduced two neural units, \textit{i.e.}, the parameterized filter and parameterized gate, to incorporate aspect information into CNN. Notably, in CNN-based methods, it is common to employ the average of aspect embeddings as the aspect representation, which would cause the loss of sequence information. To address this issue, Li \textit{et al.} \cite{li2018transformation} introduced a target-specific transformation component based on CNNs to better learn the target-specific representation.

However, due to the challenge of multiple aspects with different polarities in a sentence, context-based models usually confuse the connections between aspects and related opinion words. To this end, most recent efforts focus on leveraging the syntactic structure of the sentence to effectively establish the connection \cite{zhang-etal-2019-aspect,wang2020relational,hou-etal-2021-selective,hou-etal-2021-graph,li-etal-2021-dual-graph,pang-etal-2021-dynamic,tang2020dependency}. In practice, syntactic dependency trees are introduced to represent the sentence, and then GNNs are used to model the dependency trees and encode the syntactic information. Zhang \textit{et al.} \cite{zhang-etal-2019-aspect} first utilized dependency trees to represent sentences and then proposed graph convolution networks (GCNs) to exploit syntactical information from dependency trees. Additionally, to better connect the aspect and opinion words syntactically, Wang \textit{et al.} \cite{wang2020relational} presented a novel aspect-oriented dependency tree structure and employed a relational graph attention network to encode the tree structure. In addition, regarding sentences that have no remarkable syntactic structure, Pang \textit{et al.} \cite{pang-etal-2021-dynamic} introduced a multichannel GCN to optimally fuse syntactic and semantic information and their combinations simultaneously. Similarly, in the study of Li~\textit{et al.} \cite{li-etal-2021-dual-graph}, a dual GCN model that consists of SemGCN and SynGCN modules was used to take advantage of the complementarity of syntax structure and semantic correlations.

\subsection{Incorporating External Knowledge}
Since linguistic and commonsense knowledge can be beneficial to understanding natural language, incorporating this knowledge into deep learning models has become an active topic in many fields \cite{yasunaga2021qa, liu2018entity, zhang-etal-2019-ernie, liu2020k, xing-etal-2021-km}. A case in point is the ERNIE \cite{zhang-etal-2019-ernie}, which employed the large-scale corpora and knowledge graphs to train a knowledge-enhanced pretraining language model. ERNIE experimentally achieves great performance on various knowledge-driven downstream tasks.

However, in the task of ABSA, the existing methods fall short in exploring the knowledge to augment the sentiment analysis. One main reason for this is that the above knowledge is not explicitly expressed in the ABSA datasets. Therefore, some recent studies attempt to incorporate external knowledge to alleviate this issue \cite{ma2018targeted,wu2019aspect,zhou2020sk, zhao2021knowledge,liu2021unified,wang-tkde2021,islam2022ar}. Wu \textit{et al.} \cite{wu2019aspect} proposed a unified model to integrate sentiment and structure knowledge with contextual representations for better performance. Zhou \textit{et al.} \cite{zhou2020sk} proposed jointly encoding syntactic information and external commonsense knowledge, where the knowledge was sampled via the individual nodes. Moreover, in the study of Xing~\textit{et al.} \cite{xing-etal-2021-km}, a knowledge-enhanced BERT was introduced to obtain representations enhanced with sentiment domain knowledge to improve ABSA performance. 

Following this line of research, we introduce knowledge graphs to explicitly provide external knowledge for ABSA. \zqh{This idea is relatively similar to AR-BERT\cite{islam2022ar}, which incorporates information on aspect-aspect relations in knowledge graphs to improve the performance of existing ABSA models. While AR-BERT~\cite{islam2022ar} can achieve encouraging performance with the help of a large-scale knowledge graph, its main focus is on modeling aspect relations (captured by a complex method) from large knowledge graphs.} In contrast, we start from the multiview learning perspective and propose a novel ABSA model that uses a simpler and more efficient strategy to model knowledge graphs. Additionally, instead of only integrating external knowledge with contextual or syntactic information, we synergistically combine the knowledge with both contextual and syntactic information to obtain richer feature representations and effectively boost the performance of sentiment analysis.

\section{Knowledge Graph Augmented Network} \label{sec:method}
\subsection{Problem Formulation}
In this section, we first define the task of ABSA mathematically. Suppose we have a sentence-aspect pair $\{S, T\}$, where $S=\{w_1,w_2,...w_{start},...,w_m\}$ denotes the \textit{m}-words sentence, and $T=\{w_{start},w_{start+1},...,w_{start+n-1}\}$ denotes the \textit{n}-words aspect that is usually the subsequence of the sentence $S$. Note that $start$ is the starting index of $T$ in $S$. The goal of ABSA is to predict the sentiment polarity $y \in \{0,1,2\}$ of the sentence $S$ towards the aspect $T$, where 0, 1, and 2 denote the \textit{positive}, \textit{neutral} and \textit{negative} sentiment polarities, respectively.

\subsection{Overview of the KGAN model}
The architecture of our proposed KGAN model is shown in Fig.~\ref{Fig0}, where KGAN contains three branches, \textit{i.e}., context-, syntax- and knowledge-based branches, which learn the feature representations from multiple views. Specifically, the contextual and syntactic branches extract the contextual and syntactic features from the sentence represented by the pretrained word embeddings and explicitly establish the relevance between aspects and opinion words in the sentence. We then present the knowledge branch to model the introduced knowledge graphs and incorporate the external knowledge into the learned semantic features. In practice, knowledge graphs are first embedded into distributed representations, and a soft attention mechanism is then utilized to learn aspect-specific knowledge representations. Last, we synergistically fuse the learned multiview representations with a hierarchical fusion module.

\subsection{Multi-View Representation Learning}

\subsubsection{Context-based Representations}
Recent works~\cite{ding2020context,yang2021context} have shown that context-aware representation could successfully improve the language understanding ability, thus achieving better performance. Given the sentence-aspect pair $\{S, T\}$, we employ the popular pretrained word embedding model to represent each word of $S$ and $T$, respectively. In particular, we embed each word $w_i$ into a low-dimensional vector space with embedding matrix $E \in \mathbb{R}^{|V| \times d_w}$, where $|V|$ and $d_w$ are the size of the vocabulary and the dimension of word embeddings, respectively. The embedding matrix is usually initialized with the embeddings of pretrained models, \textit{e.g.}, the static word embedding model GloVe~\cite{pennington2014glove}. \zqh{Notably, inspired by the success of large-scale pretrained language models, \textit{e.g.}, BERT~\cite{devlin2018bert} and RoBERTa~\cite{liu2019roberta}, we can also follow~\cite{wang2020relational, tang2020dependency} and use the BERT/RoBERTa as the word embedding extractor that can be also referred to as a more sophisticated ``embedding matrix''. In practice, $S$ and $T$ are fed into the BERT/RoBERTa and the token embeddings of the last hidden layer are used as the mapped output embeddings.} Moreover, considering the benefit of positional information between context words and aspects, we follow \cite{tang-etal-2016-aspect} and encode the relative position features into the word embeddings of $S$. Thus, the sentence $S$ and aspect $T$ are converted to the corresponding word embeddings $X^s=\{x^s_1,x^s_2,...,x^s_m\}$ and $X^t=\{x^t_1,x^t_2,...,x^t_n\}$ in the end.

Based on the word embeddings, two separate bidirectional LSTMs (BiLSTMs) are used to capture the statistical dependencies in the sentence and aspect. In particular, we denote the forward operation of the LSTM as $\overrightarrow{LSTM}$ and the backward operation as $\overleftarrow{LSTM}$. The hidden state vectors $h^s_i$ and $h^t_i$ can be obtained with:
\begin{align}
h^s_i &= [\overrightarrow{LSTM}(x^s_i), \overleftarrow{LSTM}(x^s_i)], & i \in \{1,m\} \\
h^t_j &= [\overrightarrow{LSTM}(x^t_j), \overleftarrow{LSTM}(x^t_j)]. & j \in \{1,n\}
\end{align}

As a result, we obtain the hidden output of BiLSTM for the sentence as $H^s_c=\{h^s_1,h^s_2,...,h^s_i,...,h^s_m\}$, which preserves the contextual information, and the target representation as $H^t_c=\{h^t_1,h^t_2,...,h^t_n\}$. In addition to $H^s_c$, two attention mechanisms are introduced to capture the aspect-specific contextual features. Specifically, a self-attention mechanism is first used to fully learn the long-range dependencies of context. Then, we make the other soft attention mechanism assign the weight for each word of $S$ towards the $T$ and obtain the weighted aggregation as the aspect-specific contextual representation, namely, $R_c$.

\begin{figure}[t]
	\includegraphics[width=0.4\textwidth]{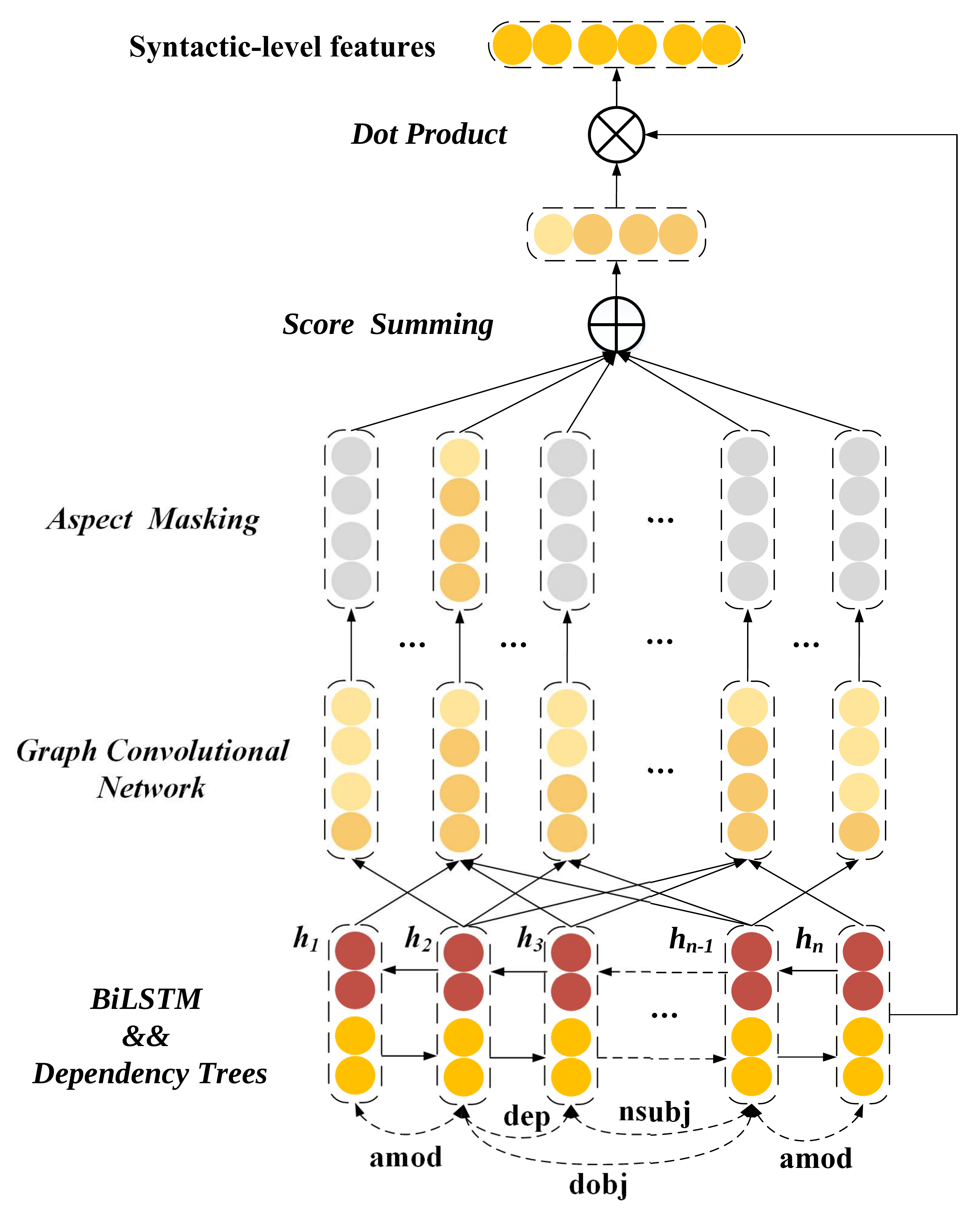} 
	\caption{Detailed illustration of the syntactic branch.}
	\label{Fig3-2}
\end{figure}

\subsubsection{Syntax-based Representations}
In the syntactic branch, we aim to leverage the explicit syntactic information to encourage the model to learn the syntax-aware representations, denoted as $R_s$, which has been shown to be helpful for many NLP tasks, \textit{e.g.}, machine translation~\cite{bastings-etal-2017-graph,ding2019recurrent}. In practice, the same pretrained word embedding model and BiLSTM are also sequentially used to obtain the hidden state vectors $H_s$. Note that we share the parameters of word embeddings and BiLSTM in both contextual and syntactic branches to reduce the computation and lighten the model size, \textit{i.e.}, $H_s=H^s_c$. Following the representative work in \cite{zhang-etal-2019-aspect}, we then employ a two-layer GCN module to extract the syntactic features of the sentence. To enable a close look at this, we show an illustration of the syntactic branch in Fig.~\ref{Fig3-2}.

First, we construct the syntactic dependency tree of the $S$ with the spaCy toolkit\footnote{{https://spacy.io/}} and obtain the adjacency matrix, namely, $A$, according to the words in the sentences. In practice, we make each word adjacent to its children's nodes and itself and set the values of adjacency nodes to ones.

The GCN is further used to encode the syntactic information of $G$ into $H_s$. In our preliminary experiments, we found that a two-layer GCN performs better than a one-layer GCN, while more layers did not improve the performance, which is consistent with results reported by previous works \cite{kipf2016semi} and \cite{yao2019graph}. We claim that a one-layer GCN hardly captures the larger neighboring information, while multiple layers lead to high complexity. Therefore, the two-layer GCN is used last in our work.
Specifically, with the operation of GCNs, the update of hidden state vectors $H_s$ can be formulated as follows:
\begin{align}
{H}_{s}^{(l+1)} &=ReLU(\dfrac{A H_s^{(l)}\ W^{(l)}}{(D +1 )} +B^{(l)}),   & l \in \{0,1\}
\end{align}
where $A$ is the adjacency matrix over the dependency tree, $D$ is the degree matrix of $A$ (\textit{i.e}, $D_{ii}=\sum_j A_{ij}$), and $W^{(l)}$ and $B^{(l)}$ are the weight and bias matrices for the ($l$+1)-th GCN layer. $H_s^{(0)}$ is the initial hidden state vector $H_s$, and $H_s^{(2)}$ is the final output of the GCNs. 

Additionally, we further introduce a graph-based attention module to learn the aspect-specific $R_s$. More specifically, the attention module first performs aspect masking on the top of $H_s^{(2)}$ to mask the nonaspect words with zero. Since the above GCNs perceived the important information in the hidden aspect state, masking the other states could alleviate the effect of noise. A dot-product attention mechanism is utilized to transition the related aspect-specific features from the initial $H_s$ towards the refined aspect features $H_s^{(2)}$ and thus syntactically build the connections of aspects and related opinion words.

\begin{figure}[]
	\centering
	\includegraphics[width=0.47\textwidth]{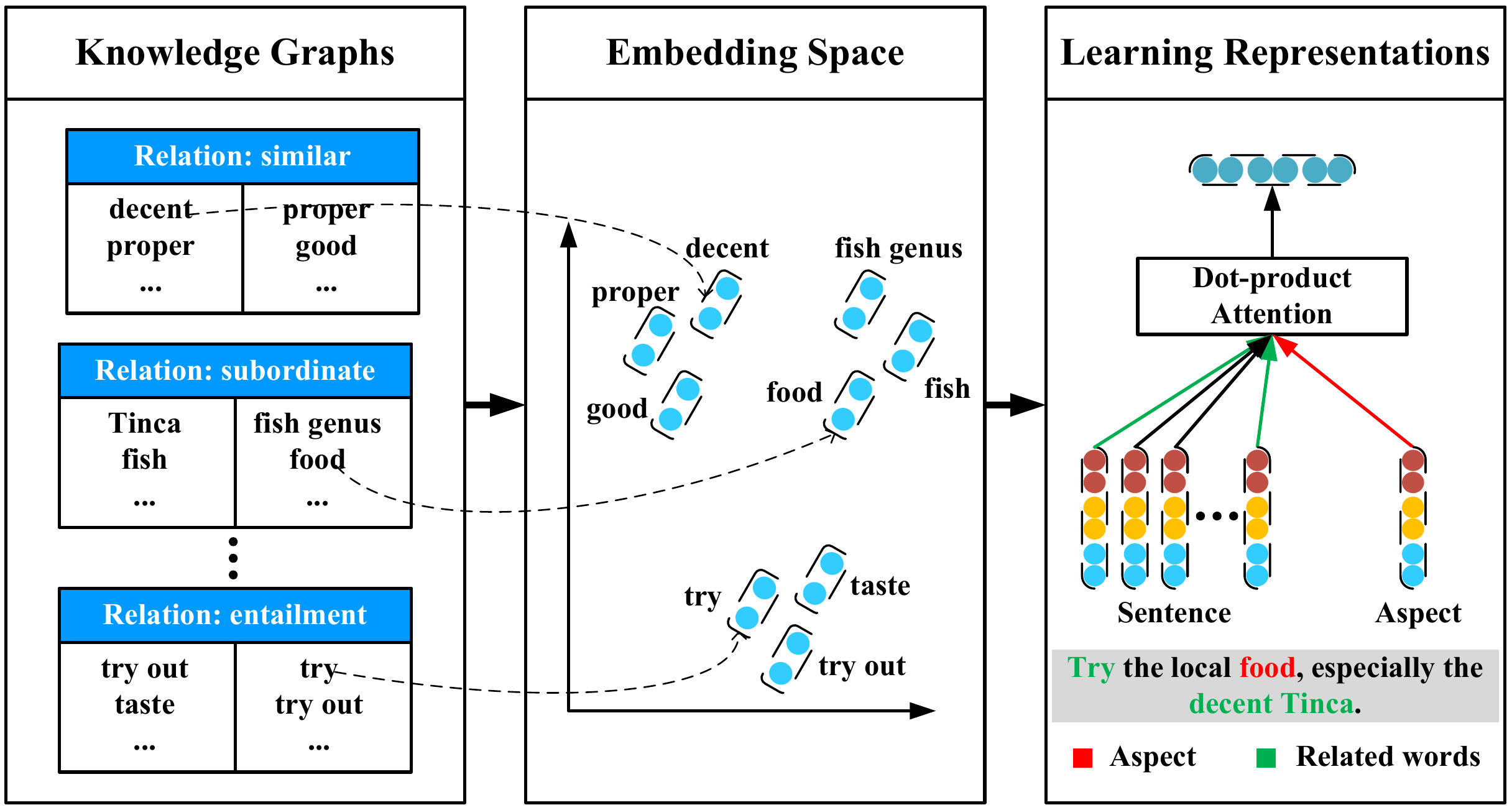} 
	\caption{The detailed illustration of the knowledge branch. In the right box, the words in red denote the aspect, while those in green are the related opinion words.}
	\label{Fig2}
\end{figure}

\subsubsection{Knowledge-based Representations.}
\label{sec.3.3.3}
To incorporate external knowledge and enrich the semantic features, we introduce the knowledge graphs of WordNet\footnote{https://wordnet.princeton.edu/}~\cite{miller1995wordnet} as the external knowledge base, which contains more than 166,000 word form and sense pairs, and employs sets of synonyms to represent the concepts. There are several semantic relations between different concepts, \textit{e.g.}, similar, opposite, part, subordinate and entailment. As with human language acquisition, we first learn the basic (simple) concepts and progressively learn the abstract (difficult) concepts. For rare difficult words, we could comprehend their meanings via their relevant normal words. Inspired by this phenomenon, we employ WordNet as prior knowledge for sentence understanding. For example, the ``Tinca" is the subordinate of the ``fish genus", which can be directly related to aspects such as ``fish" or ``food", thus alleviating the difficulty of comprehending the sentence. 

Different from Zhou~\textit{et al.} \cite{zhou2020sk}, which directly employs the graph-structure data of the knowledge base, we introduce a simple and efficient strategy to process the knowledge graphs. specifically, semantic matching approaches (see the analysis of different approaches in Sec.~\ref{sec:kge}) for the task of knowledge graph embedding (KGE) \cite{wang-tkde2017} are used to model the semantic relations of knowledge graphs into distributed representations, \textit{i.e.}, learned knowledge embeddings. \zqh{In practice, given the graph data in the form of ``entity-relation-entity'' triples, we train the entity embeddings using an open KGE toolkit, OpenKE\footnote{\zqh{https://github.com/thunlp/OpenKE.}} \cite{han2018openke}. Subsequently, we use the trained knowledge embeddings to initialize a new embedding matrix and then represent the words of $S$ and $T$ with the knowledge embedding matrix. The mapped knowledge embeddings are then concatenated with the hidden state vectors $H_s$\footnote{\zqh{Such a process can not only fuse the heterogeneous features (text and graph), but also alleviate the negative effect of sparsity and inaccuracy of knowledge embeddings.}}}. To establish the connection of $S$ and $T$ in knowledge embedding space, we further employ a soft attention mechanism to calculate the semantic relatedness of each word in $S$ and $T$ and capture the most important semantic features as aspect-specific knowledge representations, denoted as $R_k$. 
For better understanding, taking the sentence ``Try the local \textit{food}, especially the decent Tinca." and the aspect word ``food" as an example, the process of the knowledge branch is illustrated in Fig.~\ref{Fig2}. Notably, since the context word ``Tinca" is the subordinate of the aspect ``food" and they are also adjacent to each other in the knowledge embedding space, KGAN could easily capture their relatedness and make the correct prediction.

\zqh{Notably, it also should be noted that the training of these branches is not independent. Specifically, given an input sentence-aspect pair \{$S$, $T$\}, the contextual and syntactic branches apply the same embedding matrix to convert the $S$ and $T$ into the corresponding word embeddings, while the knowledge branch uses another knowledge embedding matrix to map the input entities to knowledge embeddings. As a result, through the parallel processing of three branches, KGAN can capture the aspect-specific information from multiple perspectives simultaneously.} 

\subsection{Hierarchical Fusion Module}
\begin{figure}[]

	\includegraphics[width=0.42\textwidth]{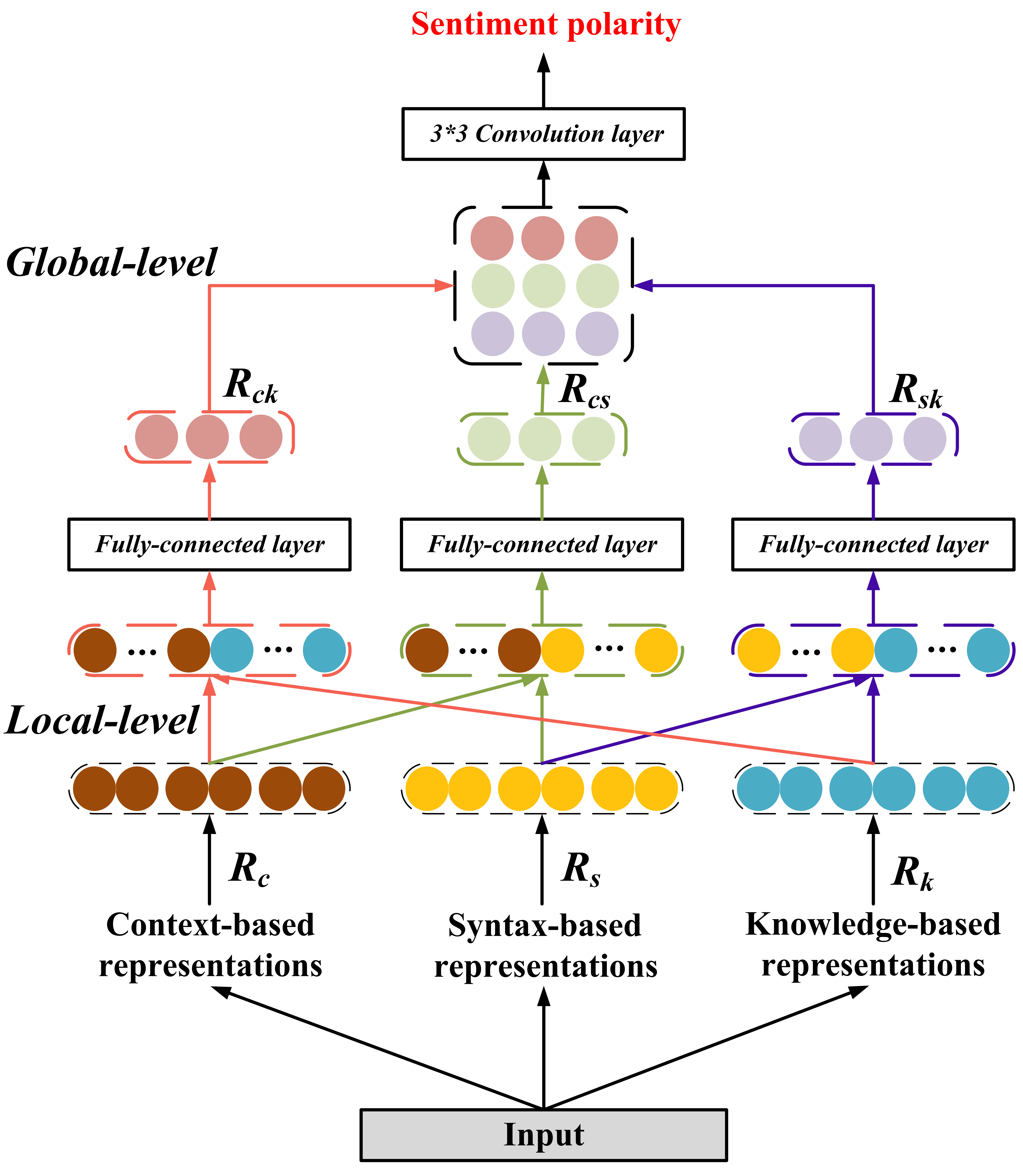} 
	\caption{Illustration of the hierarchical fusion module. }
	\label{Fig1}
\end{figure}

Since the above representations \{$R_c$,$R_s$,$R_k$\} are obtained from different views, directly fusing them may scarcely take advantage of their complementarity. To this end, we adopt a hierarchical fusion module to synergistically fuse these representations in a local-to-global manner, which could effectively boost the performance. An illustration of this fusion module is shown in Fig.~\ref{Fig1}. For ease of illustration, we employ the ``input" to represent the procedures of multiple branches.

In the local fusion procedure, we first concatenate two of the three feature representations in rows, \textit{i.e.}, [$R_c$; $R_s$], [$R_c$; $R_k$] and [$R_s$; $R_k$], where ``;'' denotes vector concatenation operator. The fused representations are fed into three separate fully connected layers to obtain the predicted sentiment features, denoted as $R_{cs}$, $R_{ck}$ and $R_{sk}$. It is noteworthy that we do not share the parameters of these fully connected layers. Subsequently, to make full use of the complementarity between multiple sentiment features, we further fuse them at the global level. Specifically, the obtained sentiment features are concatenated in columns, \textit{i.e.}, $\begin{bmatrix} R_{cs}, R_{ck}, R_{sk} \end{bmatrix}^T$, and we feed them into a 3*3 convolution layer to selectively incorporate these features.

Through the above local and global fusion procedures, we can make the feature representations benefit from each other step by step. In this way, external knowledge could be better integrated with contextual and syntactic information, thus achieving more promising performance.

Last, we cast the output of the convolution layer as the final sentiment prediction, namely, $p$, and employ the following cross-entropy loss function to guide the optimization and training. 
\begin{equation}
\mathcal{L}=-\sum_i \sum_j  y^j_i \log( p^j_i),
\end{equation}
where $i$ indexes the instance of the ABSA dataset, and $j$ indexes the sentiment polarity.

\begin{table}[t]
	\centering
	\caption{Statistics of evaluated aspect-level datasets.}
	\label{table0}
	\begin{tabular}{ccccc}
		\Xhline{1.2pt}
		\textbf{Datasets}&\textbf{Division}  & \textbf{\#Positive} & \textbf{\#Negative} & \textbf{\#Neutral} \\
		\cmidrule(){1-5}\morecmidrules\cmidrule(){1-5}
		\multirow{2}{*}{Laptop14} & Train   & 980       & 858       & 454      \\ 
		& Test    & 340       & 128       & 171      \\
		\midrule
		\multirow{2}{*}{Restaurant14}   & Train   & 2159      & 800       & 632      \\
		& Test    & 730       & 195       & 196      \\
		\midrule 
		\multirow{2}{*}{Twitter}  & Train   & 1567      & 1563      & 3127     \\
		& Test    & 174       & 174       & 346    \\ \midrule
		\multirow{2}{*}{\zqh{Restaurant15}}   & \zqh{Train}   & \zqh{912}      & \zqh{256}       & \zqh{36}      \\ 
		& \zqh{Test}    & \zqh{326}       & \zqh{182}       & \zqh{34}      \\ \midrule
		\multirow{2}{*}{\zqh{Restaurant16}}   & \zqh{Train}   & \zqh{1240}     & \zqh{439}       & \zqh{69}      \\
		& \zqh{Test}    & \zqh{469}       & \zqh{117}       & \zqh{30}      \\
		\Xhline{1.2pt}
	\end{tabular}
\end{table}

\section{Experiments} \label{sec:experiment}
\subsection{Datasets and Experimental Settings}
\zqh{Experiments are conducted on five public standard aspect-level datasets, \textit{i.e.}, Laptop14, Restaurant14, Twitter, Restaurant15 and Restaurant16. The Laptop14 and Restaurant14 datasets are from the SemEval2014 ABSA challenge \cite{pontiki-etal-2014-semeval}, Restaurant15 and Restaurant16 are from SemEval2015~\cite{pontiki2015semeval} and SemEval2016~\cite{pontiki2016semeval} challenges respectively}. The Twitter dataset is a collection of tweets \cite{dong2014adaptive}. Following \cite{tang2019progressive}, we remove a few instances with conflict sentiment polarity and list the final statistics of these datasets in Tab.~\ref{table1}. Note that we evaluate the performance with respect to Accuracy (``\textit{Acc}'') and Macro-F1 (``\textit{F1}'').

In our implementation, we carefully validate the effectiveness of our KGAN based on three pretrained models, including GloVe\footnote{http://nlp.stanford.edu/data/glove.840B.300d.zip}~\cite{pennington2014glove} with 300 dimensions, BERT~\cite{devlin2018bert} and RoBERTa\footnote{We obtain the models from https://huggingface.co/models.}~\cite{liu2019roberta}. \zqh{We use the BERT-base-uncased and RoBERTa-base models with their default settings, \textit{i.e.}, 12 layers with 768-dimensional hidden vectors. The max sequence length for BERT/RoBERTa is 512. In particular, we employ them to convert the sentence and aspect into word embeddings. Following the existing work~\cite{zhang2019ernie}, we keep the knowledge embeddings fixed during training.} The learning rates are empirically set as 1e-3 for GloVe-based KGAN, 5e-5 for KGAN-BERT and 3e-5 for KGAN-RoBERTa. \zqh{We set the batch size to 32 for most datasets, except 64 for Restaurant14\footnote{In preliminary experiments, we performed a grid search for the learning rate with \{1e-3, 1e-4, 5e-5, 3e-5, 1e-5\}, \zqh{for the batch size with \{16, 32, 64, 128\}. The reported settings are selected for \textit{best practice}.}}.} To avoid overfitting, we apply dropout on the word embeddings with a drop rate of 0.5. Adam \cite{kingma2015adam} is employed to fulfill the optimization and training. \zqh{The random seed is set to 14 in all experiments.}

\input{main_result.tex}

For comparison, we report other competitive approaches on different pretrained embeddings.
Specifically, the GloVe-based methods can be roughly divided into three categories:
\begin{enumerate}
    \item \textbf{Context-based methods:}
		\begin{itemize}
			\item \textbf{ATAE-LSTM} \cite{wang2016attention}: The aspect embedding and attention mechanism are utilized in the LSTM for aspect-level sentiment classification.
			\item \textbf{RAM} \cite{chen2017recurrent}: This method employs multiple attention and memory networks to capture the aspect-specific sentence representation. 
			\item \textbf{TNet-AS} \cite{li2018transformation}: Using the CNN as the feature extractor, TNet-AS introduces a target-specific transformation component to better incorporate the target information into the representation.
			\item \textbf{MGAN} \cite{sun-etal-2019-aspect}: This network proposes to employ the coarse-grained aspect category classification task to enhance the fine-grained aspect term classification task and introduces a novel attention mechanism to align the features between different tasks.
			\item \zqh{\textbf{MCRF-SA} \cite{xu-etal-2020-aspect}: Considering the importance of opinion spans, the network presents neat and effective multiple CRFs to extract the aspect-specific opinion spans.}
		\end{itemize}
	
	   \item \textbf{Syntax-based methods:}
		\begin{itemize}
			\item \zqh{\textbf{ASGCN} \& \textbf{ASCNN} \cite{zhang-etal-2019-aspect}: ASGCN is the first work to represent sentences with dependency trees and adopt GCN to explore the syntactical information from the dependency trees for ABSA. ASCNN is a variant of ASGCN that replaces the 2-layer GCN with 2-layer CNN.}
			\item \textbf{R-GAT} \cite{wang2020relational}: To better model syntax information, R-GAT proposes a novel aspect-oriented dependency tree structure to reshape and prune ordinary dependency parse trees.
			\item \textbf{DGEDT} \cite{tang2020dependency}: The transformer is introduced into the network to diminish the error induced by incorrect dependency trees, thus boosting the performance.
			\item \textbf{RGAT} \cite{9276424}: A relational graph attention network is proposed to make full use of the dependency label information, which is intuitively useful for the ABSA task.
			\item \zqh{\textbf{kumaGCN} \cite{chen2020inducing}: Regarding the problem of unsatisfactory dependency trees, kumaGCN associates dependency trees with induced latent aspect-specific graphs, and proposes gating mechanisms to dynamically combine information.}
			\item \textbf{DM-GCN} \cite{pang-etal-2021-dynamic}: Considering the lack of syntactic information in some bad cases, a dynamic and multichannel GCN is used to jointly model the syntactic and semantic structures for richer feature representations.
			\item \textbf{DualGCN} \cite{li-etal-2021-dual-graph}: For the problem of inaccurate dependency parsing results, Dual-GCN leverages the additional semantic information to complement the syntactic structure.
		\end{itemize}
			
\item \textbf{External knowledge-based methods:}
		\begin{itemize}
			\item \textbf{Sentic-LSTM} \cite{ma2018sentic}: To explicitly leverage commonsense knowledge, this method proposes an extension of LSTM that can use the knowledge to control the information. 
			\item \textbf{MTKEN} \cite{wu2019aspect}: Multiple sources of knowledge, \textit{i.e.}, structure and sentiment knowledge, are fused in a unified model to boost the performance.
			\item \textbf{SK-GCN} \cite{zhou2020sk}: A syntax- and knowledge-based GCN model is proposed to effectively incorporate syntactic information and commonsense knowledge by jointly modeling the dependency tree and knowledge graph.
			\item \zqh{\textbf{Sentic GCN} \cite{liang2022aspect}: This method constructs GCN via integrating the effective knowledge from SenticNet to enhance the dependency graphs of sentences.}
		\end{itemize}	
		
	\end{enumerate}
Additionally, we compare KGAN to some powerful BERT- and RoBERTa-based methods to investigate the complementarity of our KGAN with powerful pretrained language models. \zqh{Notably, despite our efforts to retrieve existing RoBERTa-based models, few models are applied to Restaurant15 and Restaurant16 datasets.}

\subsection{Main Results and Analysis}
Tab.~\ref{table1} lists the results of previous competitive models. First, in the GloVe settings, we find that our KGAN model outperforms the other cutting-edge methods on most evaluated datasets. Specifically, KGAN performs better than Sentic-LSTM and SK-GCN, which only combine external knowledge with single contextual or syntactic information, indicating the superiority of multiview representation learning. Additionally, in the Laptop14 dataset, compared to the current SOTA model DualGCN (\textit{Acc: 78.48\%; F1: 74.74\%}), KGAN (\textit{Acc: 78.91\%; F1: 75.21\%}) achieves performance improvements of 0.43\% and 0.47\% in terms of accuracy and macro-F1 score, respectively. Although the performance of KGAN (\textit{Acc: 84.46\%; F1: 77.47\%}) for Restaurant14 is suboptimal, it also outperforms all models by at least 0.19\% with respect to the accuracy metric and outperforms most models by over 1.39\% in terms of macro-F1 score. These results can demonstrate the effectiveness and superiority of our KGAN.

Interestingly, we can find that the averaged performance of context-based models is worse than their syntax-based counterparts, especially on the Restaurant14 benchmark. One possible reason for this is that the ratio of multiaspect instances in Restaurant14 (26.58\%) is higher than that of Laptop14 (20.05\%)~\cite{jiang2019challenge}, where the nonlocal modeling ability provided by the syntactic dependent trees could effectively address such a multiaspect problem. More specifically, in the group of context-based methods, the CNN-based models, \textit{e.g.}, TNet-AS, significantly outperform the LSTM-based models, \textit{e.g.}, ATAE-LSTM and RAM, on the Twitter dataset. This is because instances of the Twitter dataset are almost ungrammatical and noisy \cite{li2018transformation}, which greatly hinders the effectiveness of the LSTM.

Last, we see that the performance of our KGAN on pretrained language models, \textit{i.e.}, BERT and RoBERTa, could achieve significant and consistent improvements compared with the results for GloVe, showing the complementarity between our approach and powerful pretrained language models. Encouragingly, our KGAN on RoBERTa achieves the new SOTA among all benchmarks, while the KGAN on BERT also outperforms most cutting-edge models in the same setting.

\subsection{Ablation Study}
In this section, we conduct extensive ablation studies to investigate the effects of multiple representations and the proposed fusion module in KGAN. Additionally, we analyze the influences of different knowledge graph embedding approaches. 
Unless otherwise stated, all mentioned KGAN models below are based on GloVe, and we believe that KGAN-BERT and KGAN-RoBERTa show similar trends.

\begin{table}[]
	\centering
	\caption{Experimental results (\%) of different combinations of multi-view representations ($R_c$, $R_s$ and $R_k$ mean context, syntax, and knowledge, respectively) on Restaurant14 and Twitter datasets.
	}
	\label{table2}
	\begin{tabular}{ccccccc}
		\Xhline{1.2pt} 
    \multirow{2}{*}{\textbf{$R_c$}} & \multirow{2}{*}{\textbf{$R_s$}} & \multirow{2}{*}{\textbf{$R_k$}}   & \multicolumn{2}{c}{\textbf{Restaurant14}}  & \multicolumn{2}{c}{\textbf{Twitter}}        \\ \cmidrule(){4-7}
	&&  &Acc. (\%)         & F1 (\%)          & Acc. (\%)         & F1 (\%)                 \\  \cmidrule(){1-7}\morecmidrules\cmidrule(){1-7}   
		\checkmark &&           &81.94    &73.92   & \zqh{76.53}   &  \zqh{75.35}  \\
		&\checkmark&           &81.42    &72.85    & \zqh{74.29}  &  \zqh{73.03}      \\
		&&\checkmark             &\zqh{82.55}	&\zqh{75.07}	&\zqh{76.22}	&\zqh{74.95}     \\
		\checkmark&\checkmark &      &82.81      &75.00  &\zqh{76.85}    & \zqh{76.04}   \\
		\checkmark& &  \checkmark         &\zqh{82.92}      &\zqh{75.21}    &\zqh{77.27}   &\zqh{76.04}   \\
		&\checkmark& \checkmark        &\zqh{83.35}     &\zqh{75.25}   &\zqh{76.99}   &\zqh{75.76}    \\
		\checkmark&\checkmark&\checkmark    & \textbf{\zqh{83.46}}    & \textbf{\zqh{75.67}}   &\textbf{\zqh{78.41}}    &\textbf{\zqh{77.29}} \\
		\Xhline{1.2pt}
	\end{tabular}
\end{table}

\subsubsection{Effects of Different Multiview Representation Combinations.}
Tab.~\ref{table2} lists the results of different representation combinations, \textit{i.e.}, $\{R_c,R_s,R_k\}$. For fair comparison, we concatenate different representations and feed them into a one-layer MLP classifier to achieve multi-view fusion. Note that we do not employ the proposed hierarchical fusion module to fuse the entire representation combination ``[$R_c$,$R_s$,$R_k$]''; thus, the performances of the last row in Tab.~\ref{table2} are slightly worse than the relative ones in Tab.~\ref{table1}.

Clearly, all representations from different views are of benefit to our KGAN.
\zqh{Specifically, $R_k$ performs better than $R_c$ and $R_s$, owing to the aligned features of text and graph data. As stated in Sec.\ref{sec.3.3.3}, the knowledge embeddings (graph information) are concatenated with the hidden outputs of the BiLSTM (text information).}
Notably, with the help of $R_k$, the representation without knowledge ``[$R_c$,$R_s$]'' can achieve an averaged +1.08\% improvement, showing the effectiveness of introducing knowledge to ABSA models.
Recall that the combination of multiview representations ``[$R_c$,$R_s$,$R_k$]'' performs best among all combinations, and thus remains as the default setting.

\begin{figure}[]
	\includegraphics[width=0.47\textwidth]{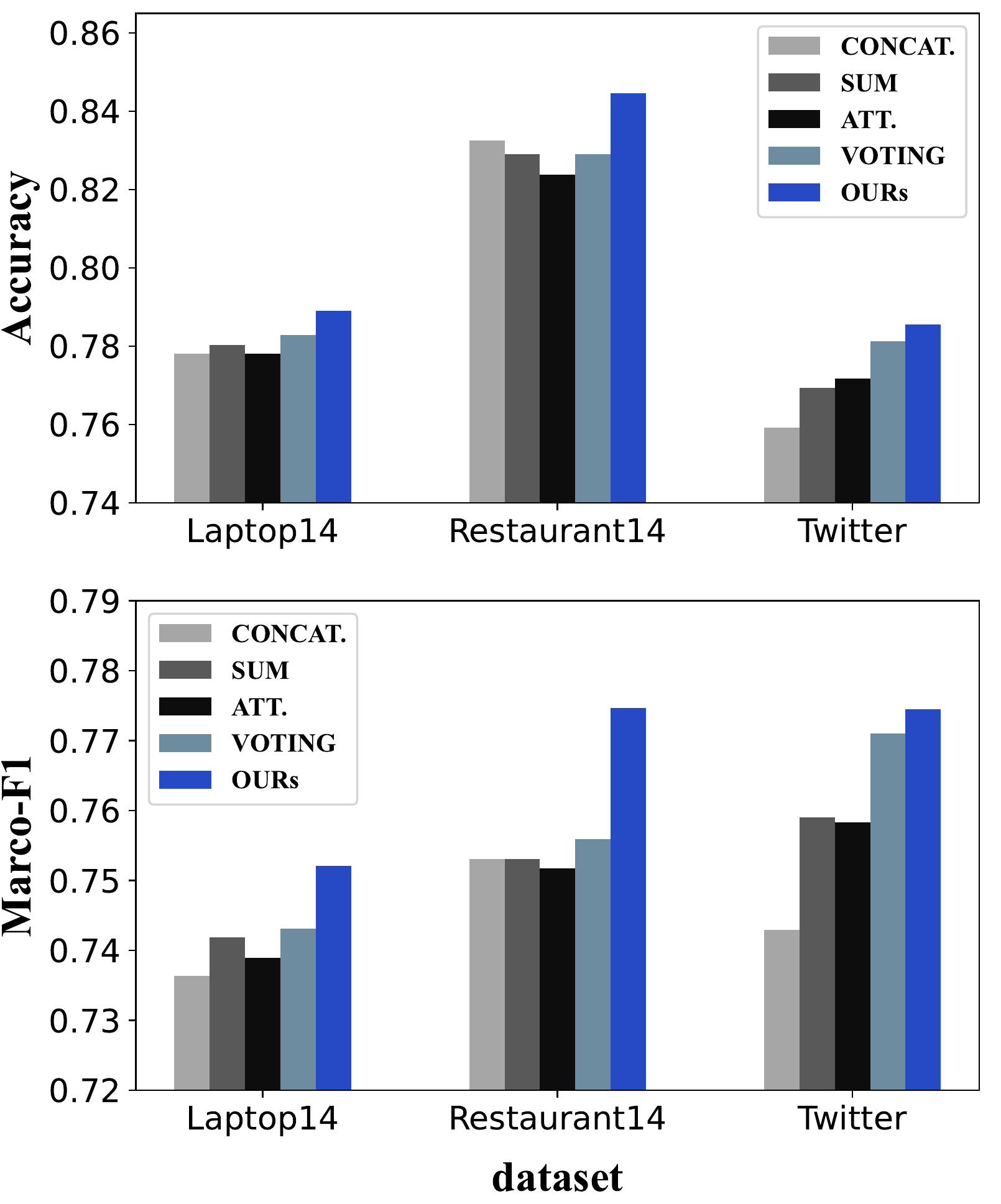} 
	\caption{The performance in terms of accuracy (upper) and macro-F1 (bottom) for different fusion strategies. \zqh{``CONCAT.", ``SUM", ``ATT." and ``VOTING'' denote concatenation strategy, elementwise summation strategy, multi-view attention mechanism strategy and voting mechanism strategy, respectively.}}
	\label{ablation1}
\end{figure}

\subsubsection{Effects of Different Fusion Strategies.}
To validate the effectiveness of our proposed hierarchical fusion module (see Fig.~\ref{Fig1}), we compare it with several typical information fusion approaches: 1) ``\textsc{Concat.}": the multiview representations are directly concatenated in rows and fused by a fully connected layer; 2) ``\textsc{Sum}": the representations are fed into three separate fully connected layers and fused via elementwise summation; \zqh{ 3) ``\textsc{Att.}'': multi-view attention mechanism is used to fuse the multi-view representations. Specifically, given the representations $\{R_c,R_s,R_k\}$, we first concatenate them to obtain the representation ``$\begin{bmatrix} R_c, R_s, R_k \end{bmatrix}^T$'' as the key and value in the attention mechanism. Sequentially, $R_c/R_s/R_k$ are used as queries to calculate the attention scores respectively. These different view scores are added together and fed into a softmax layer to obtain the final attention scores, which are then used to multiply by the value. Lastly, the fused representation is fed into an MLP layer to obtain the sentiment prediction; 4)``\textsc{Voting}'': the representations are fused by voting mechanism, \textit{i.e.}, we train an individual classifier for each view representation and perform soft voting to determine the final prediction; 
5) ``\textsc{Ours}": the representations are fused by our proposed hierarchical fusion module. }

As shown in Fig.~\ref{ablation1}, compared to the other fusion strategies, our hierarchical fusion module significantly and consistently outperforms them with respect to both accuracy and macro-F1 metrics on Laptop14 and Restaurant14 datasets. More specifically, the improvement of the macro-F1 score on Restaurant14 is 2.16\%, and the relative increase in accuracy is at least 0.63\% on Laptop14. These results provide evidence that directly fusing multiview representations (\textit{e.g.}, \textsc{Concat.} and \textsc{Sum}) is sub-optimal. In contrast, our proposed fusion module proposes to fuse these multiview representations in a \textit{local-to-global} manner, which could take full advantage of their complementarity. \zqh{Notably, it can be found that our method outperforms slightly than the \textsc{Voting} (\textit{Acc: 78.13\%; F1: 77.10\%}) on Twitter. One possible reason is that Twitter contains more rare and ungrammatical words, which are difficult to be completely covered by KGE, \textit{i.e.}, the knowledge embeddings for Twitter is relatively sparse and inaccurate, thus hindering the performance. Nevertheless, we state that when using a better KGE (\textit{e.g.}, that pretrained on larger knowledge graphs), the problem can be alleviated and the performance will be further improved (the detailed discussion can be found in Sec.~\ref{sec:kge}).}

\begin{table}[t]
	\centering
	\caption{\zqh{Analyses of knowledge embeddings from two perspectives, \textit{i.e.}, (a) different KGE approaches and (b) different knowledge graphs. Notably, the numbers in parentheses denote the embedding dimensions, and the best results are in bold.}}
	\label{table4}
	\begin{tabular}{lcccccc}
		\Xhline{1.2pt}
		\multicolumn{1}{c}{\multirow{2}{*}{\textbf{Approach}}} & \multicolumn{2}{c}{\textbf{Laptop14}}    & \multicolumn{2}{c}{\textbf{Restaurant14}} & \multicolumn{2}{c}{\zqh{\textbf{Twitter}}}     \\ \cmidrule(){2-7}
		\multicolumn{1}{c}{}  & Acc.   & F1.  & Acc.  & F1.   & \zqh{Acc.}  & \zqh{F1.}   \\ \cmidrule(){1-7}\morecmidrules\cmidrule(){1-7}
		\multicolumn{7}{c}{\zqh{\textit{(a) Analysis of different KGE approaches (WordNet is used).}}}      \\ \cmidrule(lr){1-7}
		TransE (300)     & 77.80   & 73.69   & 82.55   & 74.67  & \zqh{76.70}    & \zqh{75.54}     \\
		ComplEx     & 78.13    & 74.09   & 83.51  & 76.01    & \zqh{76.56}    & \zqh{75.41}     \\
		ANALOGY   & \textbf{78.91} & \textbf{75.21} & 83.33   & 75.19  & \zqh{76.99}    & \zqh{75.80}  \\
		DistMult  & 77.81     & 73.96    & \textbf{84.46}  & \textbf{77.47} & \zqh{\textbf{77.27}}    & \zqh{\textbf{76.17}}    \\
		\midrule \midrule
		\multicolumn{7}{c}{\zqh{\textit{(b) Analysis of different knowledge graphs (TransE is used).}}}          \\ \cmidrule(lr){1-7}
		\multicolumn{7}{l}{\zqh{WordNet (smaller)}}      \\
		\quad \zqh{-TransE (100)}    & \zqh{76.41}    & \zqh{72.11}   & \zqh{82.55}    & \zqh{74.45}    & \zqh{76.73}    & \zqh{76.74}    \\
		\quad \zqh{-TransE (50)}      & \zqh{76.88}   & \zqh{72.64}    & \zqh{82.90}    & \zqh{74.42} & \zqh{77.13}   & \zqh{75.53}   \\ \cdashline{1-7}
		\multicolumn{7}{l}{\zqh{Wikidata (larger)}}                           \\
		\quad \zqh{-TransE (100)}   & \zqh{\textbf{78.13}}  & \zqh{\textbf{73.79}}    & \zqh{\textbf{83.51}}   & \zqh{75.81}   & \zqh{77.70}    & \zqh{76.79}     \\
		\quad \zqh{-TransE (50)}     & \zqh{77.03}     & \zqh{73.10}   & \zqh{83.31}    & \zqh{\textbf{76.01}}   & \zqh{\textbf{78.55}} & \zqh{\textbf{77.45}} \\
		\Xhline{1.2pt}
		\end{tabular}
\end{table}

\subsubsection{Effects of Different Knowledge Graph Embeddings.} \label{sec:kge}
As mentioned above, we employ several simple Knowledge Graph Embedding (KGE) approaches to model the knowledge graphs into continuous embeddings. 
\zqh{To further investigate how these knowledge embeddings affect the performance of KGAN, we conduct experiments to analyze the influences of a) different KGE approaches and b) different knowledge graphs.}

\zqh{For a), we use four typical approaches to embed the same knowledge graph (WordNet)}: a translated-based method \textbf{TransE} \cite{bordes2013translating} and three semantic matching methods, \textit{i.e.}, \textbf{ComplEx} \cite{trouillon2016complex}, \textbf{ANALOGY} \cite{liu2017analogical} and \textbf{DistMult} \cite{yang2015embedding}. Tab.~\ref{table4} (a) reports the experimental results.
It can be found that the translated-based method TransE performs worse than the other semantic matching methods. Since TransE only focuses on encoding the relation information of entities, instead of semantic information, the knowledge embeddings learned by TransE fall short in enriching the semantic features of KGAN. Correspondingly, the models that could capture relational semantics are able to achieve better performance, especially ANALOGY (\textit{Acc: 78.91\%; F1: 75.21\%}) on Laptop14. More interestingly, we find that the performance of DistMult is unstable. For Laptop14, the competitive DistMult method cannot achieve the optimal performance as it does for Restaurant14. One possible reason for this is that \zqh{DistMult} falls short in modeling the semantic relation of the laptop domain. More potential reasons will be explored in future works.

\zqh{To verify b), we introduce another larger knowledge graph, \textit{i.e.}, Wikidata, which contains more than 20M entities and 594 relations. Due to the limited amount of computing resources, instead of training on Wikidata by ourselves, we use the publicly released Wikidata knowledge embeddings\footnote{\zqh{http://openke.thunlp.org/download/wikidata}} pretrained with OpenKE toolkit. Notably, only two different dimension versions (50 and 100) of Wikidata knowledge embeddings trained with TransE are provided by OpenKE. For fair comparison, we use the same TransE to train the knowledge embeddings of WordNet and also set the embedding dimension as 50 and 100. The detailed results are listed in Tab.~\ref{table4} (b) as well.}

\zqh{When using the larger and higher-quality knowledge graphs, our KGAN consistently achieves better performance among all datasets. More specifically, the performance improvement on Twitter is up to 1.42\% and 1.92\% in terms of accuracy and  macro-F1 score respectively. These results demonstrate that the larger and higher-quality knowledge graphs can further boost the performance of our KGAN. Moreover, it is noteworthy that the performance of Wikidata embeddings on Laptop14 and Restaurant14 is slightly worse than those WordNet embeddings trained with more sophisticated KGE approaches (\textit{e.g.}, ANALOGY and DistMult) in Tab.~\ref{table4} (a). We believe that our KGAN can be more advanced potentially, as the knowledge embeddings trained on large-scale knowledge graphs via more powerful KGE methods can bring more improvements\footnote{\zqh{Since it is not our focus in this work and it is time-costly to pretrain the embeddings on such a large-scale knowledge graph, we do not conduct further experiments to empirically prove it.}}.}

\begin{figure}[tp]
	\centering
	\includegraphics[width=0.40\textwidth]{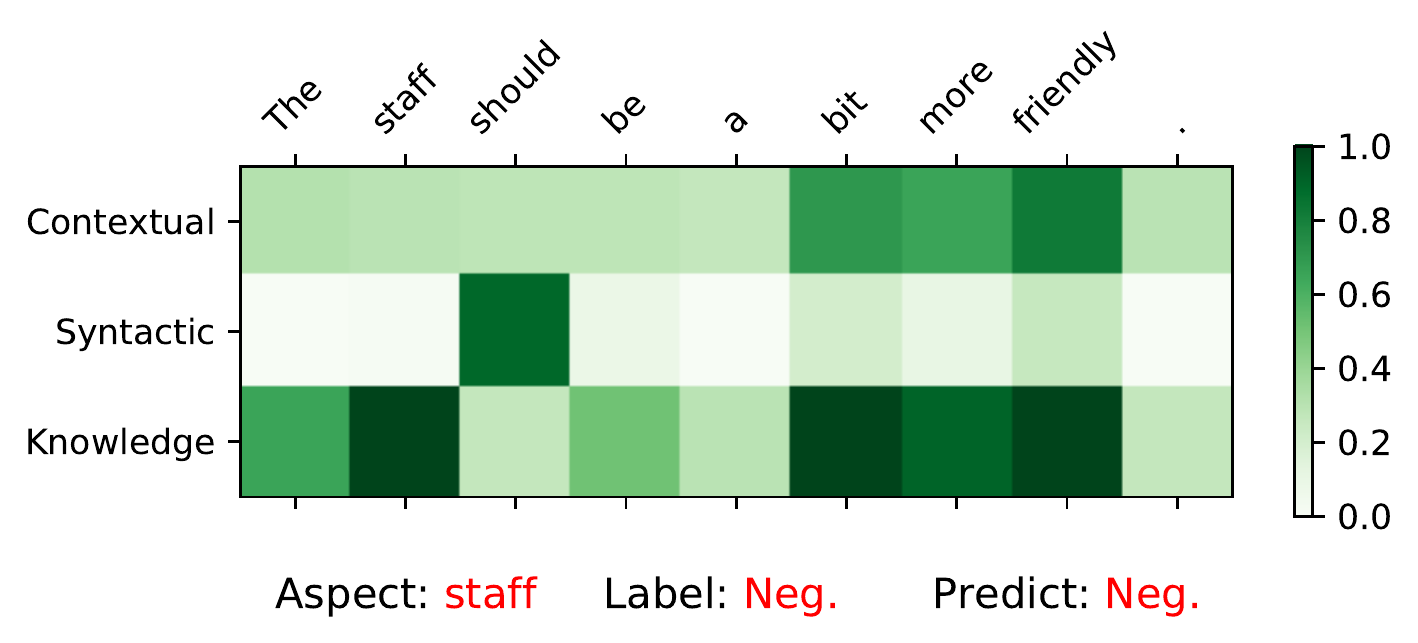} 
	\caption{Attention scores of different branches, where the aspect is \emph{\textbf{staff}}, and the label and prediction are the same \emph{\textbf{negative}}.}
	\label{fig2}
\end{figure}

\begin{figure}[tp]
	\centering
	\includegraphics[width=0.47\textwidth]{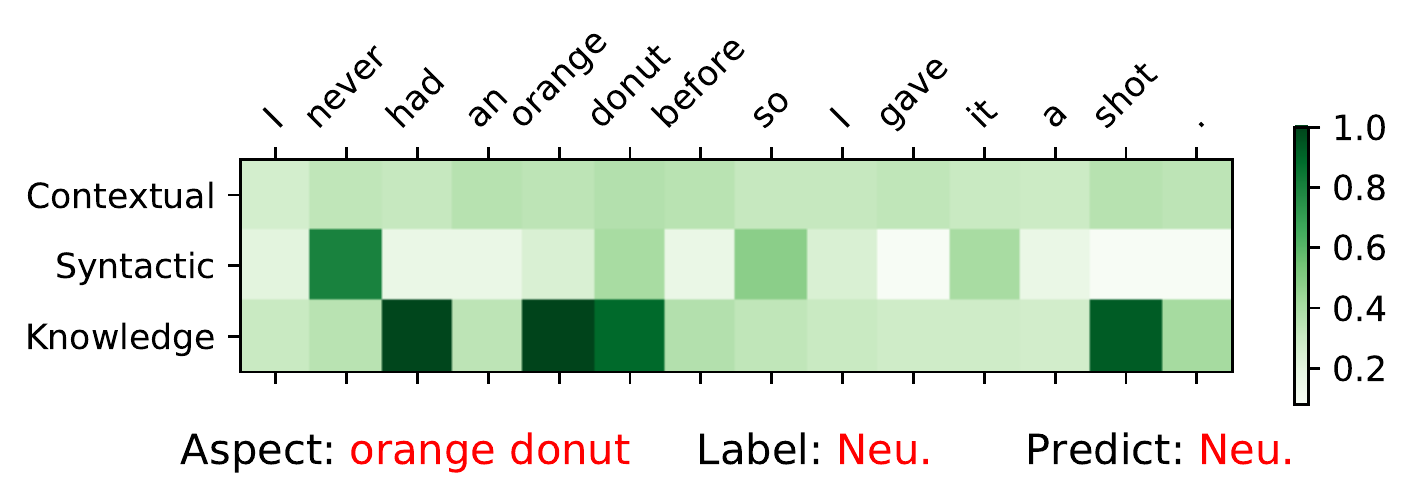} 
	\caption{In this case, the aspect is \emph{\textbf{orange donut}}, and the label and prediction are the same \emph{\textbf{neural}}.}
	\label{fig3}
\end{figure}

\begin{table*}[tp]
	\centering
	\caption{The words enclosed in [] are aspect terms, and \textit{p}, \textit{n}, and \textit{o} denote the true polarities. P, N, and O respectively denote the positive, negative and neutral predictions. The symbol \ding{55} indicates the wrong prediction.}
	\label{table5}
	\begin{tabular}{m{9cm}ccccc}
		\Xhline{1.2pt}
		\textbf{Sentences}                                                               &\textbf{RAM}        &\textbf{TNet-AS}   &\textbf{MGAN}  &\textbf{R-GAT} 	&\textbf{KGAN} \\ \cmidrule(){1-6}\morecmidrules\cmidrule(){1-6}
		1. Great $[food]_p$ but the $[service]_n$ is dreadful.                     & (P, N)     &  (P , N)    &   (P, N)    &   (P, N)      & (P, N)       \\  \midrule
		2. In mi burrito, here was nothing but dark $[chicken]_n$ that had \\that cooked last week and just warmed up in a microwave $[taste]_n$ .   & (N, P\ding{55})     &  (N , N)    &   (N, P\ding{55})    &   (N, O\ding{55})      & (N, N)    \\  \midrule
		3. $[Startup times]_n$ are incredibly long: over two minutes.              &P\ding{55}	 & N    &  N       &  N 			&N       \\   \midrule
		4. The $[folding$ $chair]_n$ I was seated at was uncomfortable.         & N   & O\ding{55}   &N    &N        & N        \\ \midrule
		5. The $[staff]_n$ should be a bit more friendly.                   &P\ding{55}  & P\ding{55}   & P\ding{55}       & P\ding{55}  &N        \\   \midrule
		6. It is really thick around the $[battery]_o$. & N\ding{55}  & N\ding{55}   & N\ding{55}      &  N\ding{55} 	 & N\ding{55}        \\   
		\Xhline{1.2pt}
	\end{tabular}
	
\end{table*}

\subsection{Case Study}
For a closer look, we select cases from several evaluated datasets for extensive case studies. We first present some instances in Tab.~\ref{table5} to show the effectiveness of our proposed KGAN. In practice, the aspects are enclosed in [], where the subscripts $p$, $n$, and ${o}$ denote the true polarities ``positive'', ``negative'', and ``neutral'', respectively. Based on the table, we can obviously find that the context-based models (\textit{e.g.}, RAM and TNet-AS) perform worse than the syntax-based method (R-GAT), which shows that the syntax-based methods could effectively encode the syntactic information, thus better establishing the connections between aspects and related opinion words. 

Moreover, our KGAN makes the correct predictions in most cases, indicating its effectiveness and superiority. In particular, the second sentence demonstrates that our KGAN model can effectively address complicated and informal sentences with the help of multiview representation learning. However, it should also be noted that all models predict erroneously in the last case, since the models are likely to focus on the opinion word ``thick", which misleads the comprehension of this sentence. Such a phenomenon indicates that KGAN requires further improvement to understand complicated sentences.

Additionally, we select two cases to show exactly how the multiview representations affect the KGAN. For both cases, we visualize the final attention weights of multiple branches and show the results in Fig.~\ref{fig2} and Fig.\ref{fig3}, respectively. The sentences are as follows:
\begin{enumerate}
	\item The $[staff]_n$ should be a bit more friendly.
	\item I never had an $[orange \ donut]_o$ before so I gave it a shot.
\end{enumerate}

In the first case, for the aspect ``staff", both representations from contextual and knowledge views focus on the opinion words ``bit more friendly", which facilitates the KGAN to understand the sentence and make the correct prediction. Moreover, we find that the representation from the syntactic view pays more attention to the other word ``should", which is also important for comprehension. These attention results demonstrate the complementarity of multiview representations.

On the other hand, in the second instance, the representation from the contextual view hardly captures the aspect-specific contextual information, while the representation from the syntactic view also fails to focus on the closely-related opinion words. However, the representation from the knowledge view can effectively extract the important related words ``had" and ``shot". One possible reason for this is that the opinion words are not syntactically adjacent to the aspect, thus leading to the difficulty of capturing important semantic information. Instead, the ``orange donut" is much closer to ``had" in the introduced knowledge graph, which allows the knowledge branch to easily capture the relatedness. This case shows the significance of incorporating external knowledge and confirms our contribution.

\subsection{Discussion}

\begin{table}[t]
	\centering
	\caption{\zqh{Comparison of KGE-based and GCN-based KGAN on the training efficiency and performance. ``DP. (s)'' denotes the time of data processing. \textit{K}-KGAN and \textit{SK}-KGAN are two GCN-based implementations on our KGAN.}}
	\label{table6}
	\begin{tabular}{lcc|cccc}
	\Xhline{1.2pt}
	\multicolumn{1}{c}{\multirow{2}{*}{\zqh{\textbf{Method}}}} & \multicolumn{2}{c|}{\zqh{\textbf{Efficiency (avg.)}}}               & \multicolumn{2}{c}{\zqh{\textbf{Laptop14}}}    & \multicolumn{2}{c}{\zqh{\textbf{Restaurant14}}} \\ \cmidrule(){2-7}
	\multicolumn{1}{c}{}                        & \multicolumn{1}{c}{\zqh{DP. (s)}} & \multicolumn{1}{c|}{\zqh{FLOPs}} & \zqh{Acc.}           & \zqh{F1.}             & \zqh{Acc.}            & \zqh{F1.}             \\ \cmidrule(){1-7}\morecmidrules\cmidrule(){1-7}
	\zqh{KGAN (Full)}                                  & \zqh{25.12}                                & \zqh{2.64 G}                          & \zqh{\textbf{78.91}} & \zqh{\textbf{75.21}} & \zqh{\textbf{84.46}}  & \zqh{\textbf{77.47}} \\
	\zqh{\quad --[$R_c$,$R_s$]}          & \zqh{22.72}                                &  \zqh{2.60 G}                          & \zqh{77.30}           & \zqh{73.65}          & \zqh{82.81}           & \zqh{75.00}             \\ \midrule
	\zqh{\textit{K}-KGAN}                                      &  \zqh{46.26}                               &   \zqh{6.21 G}                         & \zqh{77.97}          & \zqh{73.99}          & \zqh{82.12}           & \zqh{73.76}          \\
	\zqh{\textit{SK}-KGAN}                                     &  \zqh{46.10}                               &  \zqh{6.18 G}                          & \zqh{77.66}          & \zqh{73.92}          & \zqh{82.90}            & \zqh{74.42}         \\
 	\zqh{\textit{AR}-KGAN}      &\zqh{42.93}              &\zqh{2.66 G}                         &\zqh{77.66}     & \zqh{74.16}     &\zqh{83.25}       &\zqh{75.77}       \\
	\Xhline{1.2pt}
	\end{tabular}
	\end{table}

\subsubsection{\zqh{Comparison between KGE-based and subgraph-based Models.}}
\zqh{As stated in Sec.~\ref{sec:introduction}, considering that current subgraph-based (or refer to GCN-based) methods usually require extra process to construct the aspect-specific subgraphs, which would lead to excessive computation, we attempt to integrating knowledge from another perspective, \textit{i.e.}, using KGE. Here, to verify whether KGE is more suitable for our KGAN, we compare the training efficiency and performance between KGE and current subgraph-based methods.} 

\zqh{Specifically, we follow the currently existing works~\cite{zhou2020sk,islam2022ar} and provide three implementations to model the aspect subgraphs. For the first solution (denoted as \textit{K}-KGAN), we simply replace the KGE module in the knowledge branch of KGAN with a two-layer GCN (similar to SK-GCN$_1$ \cite{zhou2020sk}), while keeping the other branches and feature fusion module unchanged. For the second method (denoted as \textit{SK}-KGAN), we merge the syntactic dependency graph and aspect-specific knowledge graph and model them jointly with the GCN module in the syntax branch (similar to SK-GCN$_2$ \cite{zhou2020sk}), \textit{i.e.}, removing the knowledge branch. For the last one (denoted as AR-KGAN), inspired by AR-BERT~\cite{islam2022ar}, we use the popular
GraphSAGE algorithm~\cite{hamilton2017inductive} to learn the aspect embeddings based on the subgraphs, which are used to replace the knowledge representation of knowledge branch in our KGAN.

Tab.~\ref{table6} shows the detailed results. Obviously, the KGE-based KGAN spends less time in data processing (smaller ``DP. (s)'') and requires less computation (smaller ``GFLOPs'') than other subgraph-based models, indicating the efficiency of KGE method. On the other hand, compared to the baseline ``[$R_c$,$R_s$]'' (without external knowledge), both KGE- and subgraph-based methods achieve better performance, continuing to confirm the significance of integrating external knowledge. To be more specific, it is noteworthy that our KGE-based KGAN achieves the best performance among all datasets, which shows its effectiveness. These results indicate that the KGE is indeed more suitable for our KGAN than current subgraph-based methods, but we also believe that it is good to keep in mind that subgraph-based methods can perform quite well in some scenarios when intrinsic relations in subgraphs can provide more valuable information both among implicit aspects and aspect-context correlations.}

\subsubsection{Latency and Model Size.}
Introducing external knowledge will admittedly increase the latency and model size~\cite{li2018transformation,wang2020relational,pang-etal-2021-dynamic}. We therefore perform a contrastive investigation on whether we achieve a good trade-off between efficiency and performance. For a fair comparison, all experiments are trained and tested on an Nvidia GTX-1660 SUPER. Tab.~\ref{table7} shows the details of KGAN and previous models, including ATAE-LSTM, R-GAT, and DM-GCN. Their corresponding performances can be found in Tab.~\ref{table1}. Clearly, although our proposed KGAN model is more complicated than the previous context-based ATAE-LSTM, we can achieve significantly better performance (averaged +10.87\%  macro-F1 scores). Compared to more recent models, \textit{i.e.}, R-GAT and DM-GCN, our KGAN could achieve better performance while preserving comparable or better latency. The main reason for this is that these models introduced the sophisticated module to capture the semantic features, leading to larger model sizes, while we only employ the original BiLSTM as the feature extractor and share its parameters between three branches, greatly decreasing the model parameters.
\textbf{Takeaway:}\textit{ our KGAN establishes a good trade-off between efficiency and performance due to our parameter-sharing mechanism.}

\begin{table}[t]
	\centering
	\caption{Comparison of latency and model size with previous works on Restaurant14. ``\#Params.(M)'': number of parameters in millions. ``Train(s)'' and ``Test(s)'' indicate the averaged training time (seconds) of each epoch, and the inference time of all testing samples, respectively.}
	\label{table7}
	\begin{tabular}{cccc}
		\Xhline{1.2pt}
		\multirow{2}{*}{\textbf{Model}} & \multicolumn{2}{c}{\textbf{Latency}}    & \multicolumn{1}{c}{\textbf{Model Size}}          \\ \cmidrule(){2-4} 
		&Train (s)         & Test (s)          &\#Params. (M)                      \\  \cmidrule(){1-4}\morecmidrules\cmidrule(){1-4}
		ATAE-LSTM         &1.98      &0.71        &2.53      \\
		R-GAT       	&4.61 	&0.88 	&3.72       \\
		DM-GCN          &7.04 	&2.25 	&8.80       \\
		KGAN (Ours)      &4.41 	&1.31 	&3.62                   \\
		\Xhline{1.2pt}
	\end{tabular}
\end{table}

\begin{figure}[]
    \centering
	\includegraphics[width=0.45\textwidth]{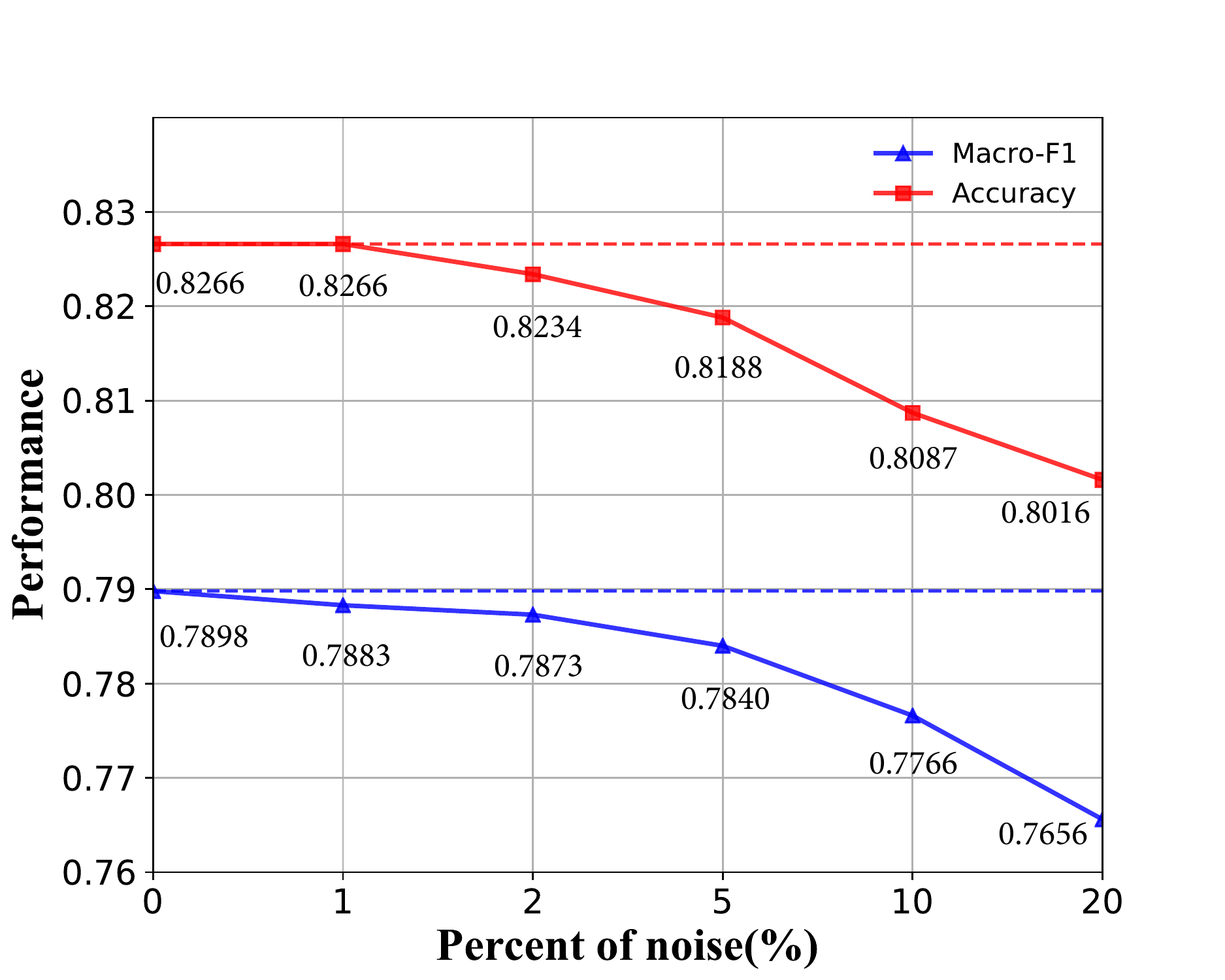} 
	\caption{Experiments with noise attacks on the KGAN-BERT model, where the results are evaluated on the Laptop14 dataset. The ratio of noisy knowledge ranges from 1\% to 20\%.}
	\label{noise}
\end{figure}

\subsubsection{Robustness of our KGAN.} 
Some researchers may doubt that the incorporated external knowledge may function as the regularizer~\cite{natarajan2013learning}. To dispel such doubt, we investigate whether our model is robust to noisy knowledge by introducing different percentages of noise to the knowledge embeddings on the Laptop14 dataset. Notably, noise attacks are widely used in the NLP community to investigate the robustness of models, such as neural machine translation \cite{edunov-etal-2018-understanding, ding-etal-2020-self}. In particular, we randomly initialize some of the knowledge embeddings as noise. As shown in Fig.~\ref{noise}, our proposed multiview representation learning method can tolerate slight noise, \textit{e.g.,} 1\%, 2\% and 5\%, and maintain performance to some extent. However, as noise increases, \textit{e.g.}, to 20\%, the performance deteriorates, indicating that noisy knowledge does not function as regularization. \textbf{Takeaway:}~\textit{1) our multiview knowledge representation approach is robust to slight noise; 2) KGAN does not benefit from noise as much as it benefits from incorporated knowledge.}

\section{Conclusion} \label{sec:conclusion}
We present a novel knowledge graph augmented network for ABSA that incorporates external knowledge to augment semantic information. Specifically, KGAN captures the sentiment features from three different views: context, syntax and knowledge. These multiview feature representations are fused via a hierarchical fusion module. Extensive experiments demonstrate the effectiveness and robustness of our proposed KGAN. Ablation experiments and case studies show the complementarity between contextual, syntactic and external knowledge, confirming our claim. Extensive analyses illustrate that our KGAN can achieve a better trade-off between latency and performance and is robust to slight noise attacks.

Future work includes validating the proposed KGAN multi-view representation approach in other challenging language understanding tasks, \textit{e.g.}, reading comprehension~\cite{zhang2020sg}, intent identification and slot filling~\cite{wu2020slotrefine}, and end-to-end language generation tasks~\cite{liu2020understanding}, \textit{e.g.}, translation, summarization and grammatical error correction tasks.

\appendices


\ifCLASSOPTIONcompsoc
  \section*{Acknowledgments}
\else
  \section*{Acknowledgment}
\fi

This work was supported in part by the National Natural Science Foundation of China under Grants 62076186 and 62225113, and in part by the Science and Technology Major Project of Hubei Province (Next-Generation AI Technologies) under Grant 2019AEA170. The numerical calculations in this paper have been done on the supercomputing system in the Supercomputing Center of Wuhan University.

\ifCLASSOPTIONcaptionsoff
  \newpage
\fi

\bibliographystyle{IEEEtran}
\bibliography{tkde-v2.bib}

\begin{IEEEbiography}[{\includegraphics[width=1in,height=1.25in,clip,keepaspectratio]{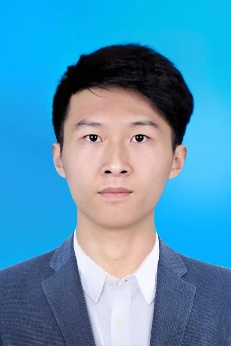}}]{Qihuang Zhong}
	is currently pursuing a Ph.D. degree in Artificial Intelligence from the School of Computer Science, Wuhan University. His research interests include language model pretraining, natural language understanding and generation. He has authored or co-authored several papers at top conferences and international journals, including IEEE TKDE, EMNLP, COLING and \textit{etc}. He won the general language understanding (GLUE) and more difficult language understanding (SuperGLUE) challenges.
\end{IEEEbiography}

\begin{IEEEbiography}[{\includegraphics[width=1in,height=1.25in,clip,keepaspectratio]{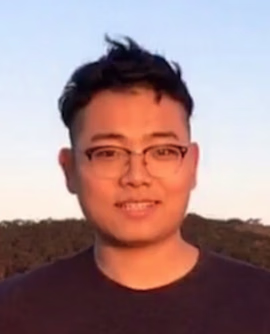}}]{Liang Ding} received Ph.D. from the University of Sydney. He is currently an algorithm scientist with JD.com and leading the NLP research group at JD Explore Academy. He works on deep learning for NLP, including language model pretraining, language understanding, generation, and translation. He published over 30 research papers in NLP/AI, including ACL, EMNLP, NAACL, COLING, ICLR, AAAI, SIGIR, and CVPR, and importantly, some of his works were successfully applied to the industry. He served as Area Chair for ACL 2022 and Session Chair for SDM 2021 and AAAI 2023. He won many AI challenges, including SuperGLUE/ GLUE, WMT2022, IWSLT 2021, WMT 2020, and WMT 2019.
\end{IEEEbiography}

\begin{IEEEbiography}[{\includegraphics[width=1in,height=1.25in,clip,keepaspectratio]{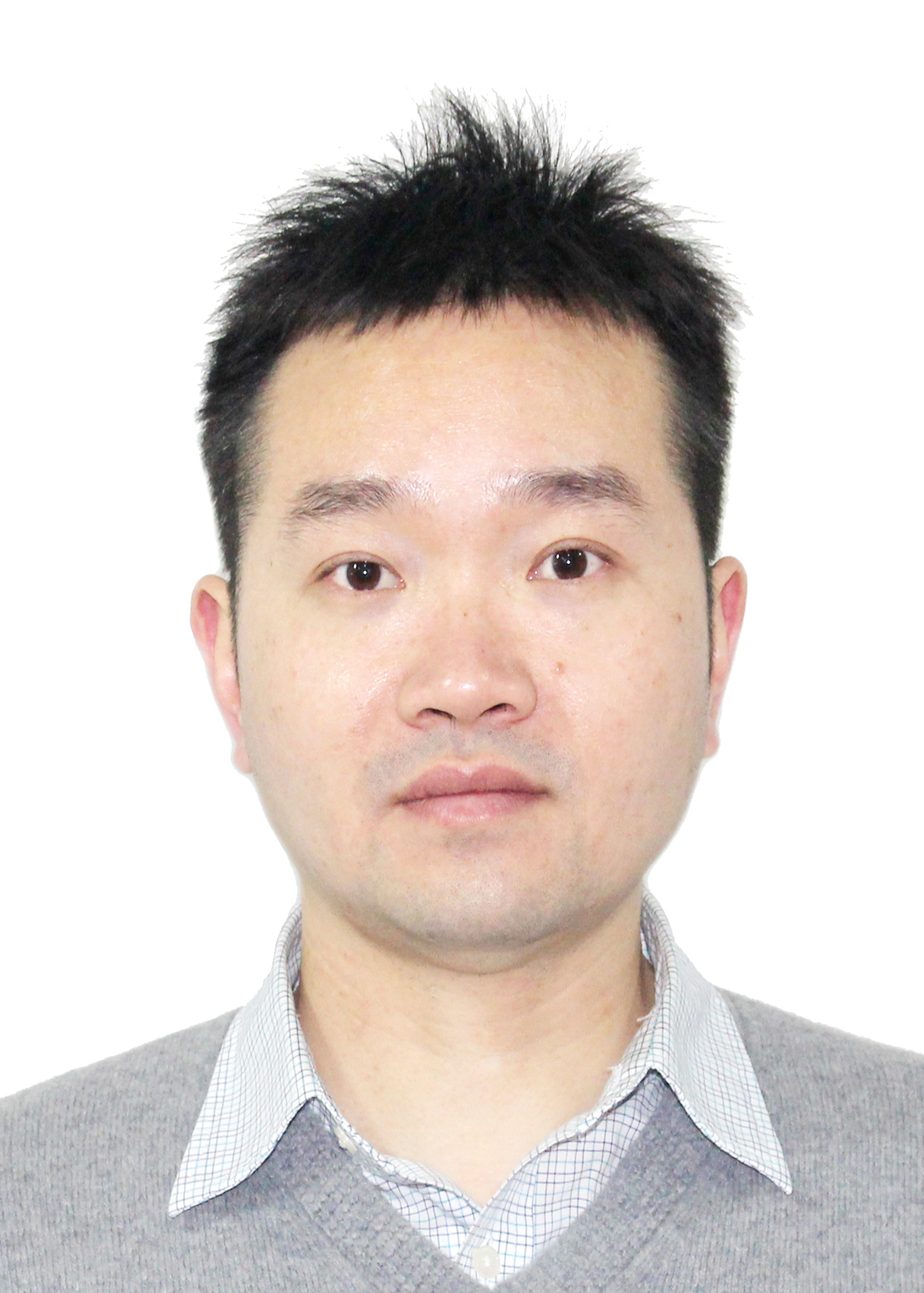}}]{Juhua Liu}
	received an M.S. and Ph.D. in Graphic Communication from the School of Printing and Packaging, Wuhan University, Wuhan, China, in 2007 and 2014, respectively. He is currently an Associate Professor with the Research Center for Graphic Communication, Printing and Packaging, and Institute of Artificial Intelligence, Wuhan University. His research interests mainly include image processing, computer vision, natural language processing and machine learning. He has published more than 30 research papers in CV/NLP/AI, including IJCV, IEEE TIP, IEEE TKDE, AAAI, IJCAI and EMNLP, \textit{etc}. He serves as a reviewer of more than 10 Science Citation Index (SCI) magazines, including IEEE TIP, IEEE TCYB, IEEE TNNLS, IEEE TIM, and regularly serves as PC member of CVPR, AAAI, IJCAI, ICASSP and ICME.
\end{IEEEbiography}

\begin{IEEEbiography}[{\includegraphics[width=1in,height=1.25in,clip,keepaspectratio]{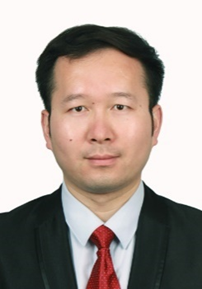}}]{Bo Du} (M'10-SM'15) received the Ph.D. degree in Photogrammetry and Remote Sensing from State Key Lab of Information Engineering in Surveying, Mapping and Remote Sensing, Wuhan University, Wuhan, China in 2010. He is currently a professor with the School of Computer Science and Institute of Artificial Intelligence, Wuhan University. He is also the director of National Engineering Research Center for Multimedia Software, Wuhan University, Wuhan, China. He has more than 80 research papers published in the IEEE Transactions on Pattern Analysis and Machine Intelligence(TPAMI), IEEE Transactions on Image Processing (TIP), IEEE Transactions on cybernetics (TCYB), IEEE Transactions on Geoscience and Remote Sensing (TGRS), \textit{etc}. Fourteen of them are ESI hot papers or highly cited papers. His major research interests include machine learning, computer vision, and image processing. He is currently a senior member of IEEE. He serves as associate editor for Neural Networks, Pattern Recognition and Neurocomputing. He also serves as a reviewer of 20 Science Citation Index (SCI) magazines, including IEEE TPAMI, TCYB, TGRS, TIP, JSTARS, and GRSL. He won the Highly Cited Researcher (2019\textbackslash2020) by the Web of Science Group. He won IEEE Geoscience and Remote Sensing Society 2020 Transactions Prize Paper Award. He won the IJCAI (International Joint Conferences on Artificial Intelligence) Distinguished Paper Prize, IEEE Data Fusion Contest Champion, and IEEE Workshop on Hyperspectral Image and Signal Processing Best paper Award, in 2018. He regularly serves as senior PC member of IJCAI and AAAI. He served as area chair for ICPR.
\end{IEEEbiography}

\begin{IEEEbiography}[{\includegraphics[width=1in,height=1.25in,clip,keepaspectratio]{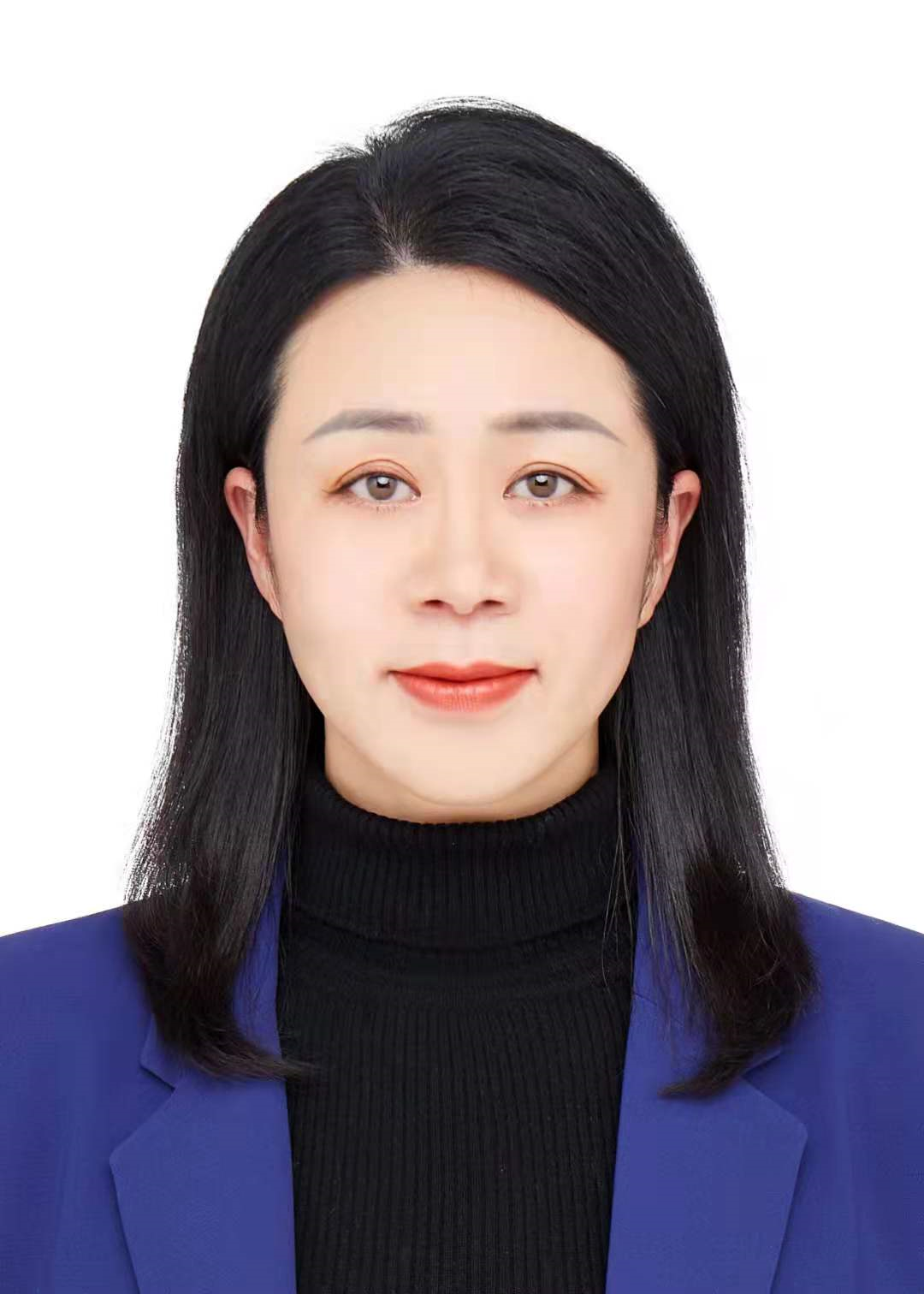}}]{Hua Jin}
	received M.D. and Ph.D. degrees from Kunming Medical University. She is a professor of the affiliated hospital of Kunming University of Science and Technology, Kunming, China. She has authored and coauthored more than 50 scientific articles. Her past and present research interests include medical big data, computer vision and perioperative analgesia.
\end{IEEEbiography}

\begin{IEEEbiography}[{\includegraphics[width=1in,height=1.25in,clip,keepaspectratio]{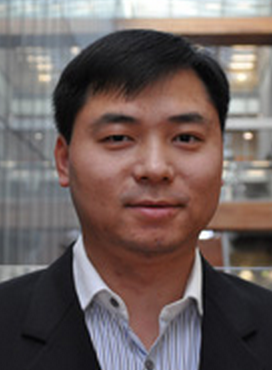}}]{Dacheng Tao}
(F'15) is currently the president of JD Explore Academy and a Senior Vice President of JD.com. He is also an advisor and chief scientist of the digital science institute at the University of Sydney. He mainly applies statistics and mathematics to artificial intelligence and data science. His research interests range across computer vision, data science, image processing, machine learning, and video surveillance. His research results have been presented in one monograph and 500+ publications in prestigious journals and at prominent conferences, such as IEEE T-PAMI, T-NNLS, T-IP, JMLR, IJCV, NIPS, ICML, CVPR, ICCV, ECCV, ICDM, and ACM SIGKDD, with several Best Paper awards, such as the Best Theory/Algorithm Paper Runner Up Award at IEEE ICDM07, the Best Student Paper Award at IEEE ICDM13, the Distinguished Student Paper Award at the 2017 IJCAI, the 2014 ICDM 10-Year Highest-Impact Paper Award, and the 2017 IEEE Signal Processing Society Best Paper Award. He was a recipient of the 2015 Australian Scopus-Eureka Prize, the 2015 ACS Gold Disruptor Award, and the 2015 UTS Vice-Chancellor's Medal for Exceptional Research. He is a Fellow of AAAS, OSA, IAPR, and SPIE.
\end{IEEEbiography}

\end{document}

%% file: main_result.tex
 \begin{table*}[t]
	\centering
	\caption{Comparison with previous work on different datasets. Most results are retrieved from corresponding papers. \zqh{``$\dag$'', ``$\ddag$'' and ``$\S$'' indicate the results are collected from \cite{zhou2020sk}, \cite{wang2020relational} and \cite{dai-etal-2021-syntax}, respectively. ``-'' means the results are not reported in their original works.}
 Notably, we report the averaged performance of KGAN with 5 random seeds to avoid stochasticity. The best results for each pretrained model are in bold, while the second-best results are underlined.}
	\label{table1}
	\begin{tabular}{llcccccccccc}
		\Xhline{1.2pt}
		\multirow{2}{*}{\textbf{Embedding}} & \multirow{2}{*}{\textbf{Method}}           & \multicolumn{2}{c}{\textbf{Laptop14}}                  & \multicolumn{2}{c}{\textbf{Restaurant14}}    & \multicolumn{2}{c}{\textbf{Twitter}}                   & \multicolumn{2}{c}{\zqh{\textbf{Restaurant15}}} & \multicolumn{2}{c}{\zqh{\textbf{Restaurant16}}} \\   \cmidrule(lr){3-12}
		&  & Acc.(\%)  & F1.(\%)  & Acc.(\%)  & F1.(\%) & Acc.(\%) & F1.(\%)  & \zqh{Acc.(\%)}  & \zqh{F1.(\%)}  & \zqh{Acc.(\%)}    & \zqh{F1.(\%)}  \\ \cmidrule(lr){1-12}\morecmidrules\cmidrule(lr){1-12}
		\multirow{18}{*}{GloVe}                                                                                  
        & ATAE-LSTM$^\dag$ \cite{wang2016attention}     & 68.88    & 63.93     & 78.60   & 67.02   & 68.64    & 66.60     & \zqh{78.48}   & \zqh{62.84}   & \zqh{83.77}   & \zqh{61.71}     \\
		& RAM \cite{chen2017recurrent}     & 74.49    & 71.35   & 80.23     & 70.80  & 69.36    & 67.30      & \zqh{79.98}   & \zqh{60.57}          & \zqh{83.88}  & \zqh{62.14}   \\
		& TNet-AS \cite{li2018transformation}      & 76.54      & 71.75   & 80.69    & 71.27    & 74.97     & 73.60    & \zqh{78.47}    & \zqh{59.47}   & \zqh{89.07}   & \zqh{70.43}    \\
		& MGAN \cite{li2019exploiting}    & 76.21  & 71.42   & 81.49     & 71.48    & 74.62     & 73.53   & \zqh{-}     & \zqh{-}     & \zqh{-}   & \zqh{-}   \\
		& \zqh{MCRF-SA \cite{xu-etal-2020-aspect}}   & \zqh{77.64}  & \zqh{74.23}     & \zqh{82.86}     & \zqh{73.78}  & \zqh{-}    & \zqh{-}     & \zqh{80.82} & \zqh{61.59}    & \zqh{89.51}   & \zqh{\textbf{75.92}}   \\ 
        \cmidrule(lr){2-12}
		& \zqh{ASCNN \cite{zhang-etal-2019-aspect}}    & \zqh{72.62}   & \zqh{66.72}     & \zqh{81.73}    & \zqh{73.10}   & \zqh{71.05}   & \zqh{69.45}      & \zqh{78.48}   & \zqh{58.90}   & \zqh{87.39} & \zqh{64.56}  \\
		& \zqh{ASGCN \cite{zhang-etal-2019-aspect}}    & \zqh{75.55}    & \zqh{71.05}     & \zqh{80.86}       & \zqh{72.19}       & \zqh{72.15}     & \zqh{70.40}    & \zqh{79.34}           & \zqh{60.78}          & \zqh{88.69}   & \zqh{66.64}   \\
		& R-GAT \cite{wang2020relational}     & 77.42   & 73.76    & 83.3    & 76.08  & 75.57        & 73.82      & \zqh{80.83}           & \zqh{64.17}          & \zqh{88.92}   & \zqh{70.89}   \\
		& DGEDT \cite{tang2020dependency}   & 76.80     & 72.30   & 83.90    & 75.10   & 74.80   & 73.40     & \zqh{82.10}           & \zqh{65.90}          & \zqh{\underline{90.80}}   & \zqh{73.80}  \\
		& RGAT \cite{9276424}      & 78.02     & 74.00   & 83.55   & 75.99 & 75.36    & 74.15       & \zqh{-}               & \zqh{-}              & \zqh{-}       & \zqh{-}       \\
		& \zqh{kumaGCN \cite{chen2020inducing}}     & \zqh{76.12}     & \zqh{72.42}   & \zqh{81.43}       & \zqh{73.64}       & \zqh{72.45}  & \zqh{70.77}   & \zqh{80.69}   & \zqh{65.99}   & \zqh{89.39}   & \zqh{73.19}    \\
		& DM-GCN \cite{pang-etal-2021-dynamic}   & \underline{78.48}    & \underline{74.90}    & 83.98     & 75.59    & \underline{76.93}   & \underline{75.90}  & \zqh{-}      & \zqh{-}   & \zqh{-}       & \zqh{-}       \\
		& DualGCN \cite{li-etal-2021-dual-graph}    & 78.48    & 74.74   & \underline{84.27} & \textbf{78.08}    & 75.92    & 74.29    & \zqh{-}      & \zqh{-}   & \zqh{-}       & \zqh{-}    \\ 
		\cmidrule(lr){2-12}
		& Sentic-LSTM$^\dag$ \cite{ma2018sentic}   & 70.88   & 67.19     & 79.43     & 70.32  & 70.66       & 67.87     & \zqh{79.55}           & \zqh{60.56}          & \zqh{83.01}   & \zqh{68.22}    \\
		& MTKEN \cite{wu2019aspect}    & 73.43    & 69.12     & 79.47     & 68.08   & 69.80        & 67.54        & \zqh{80.67}           & \zqh{58.38}          & \zqh{88.28}   & \zqh{66.15}     \\
		& SK-GCN \cite{zhou2020sk}    & 77.62     & 73.84     & 81.53    & 72.90    & 71.97   & 70.22     & \zqh{80.12}           & \zqh{60.70}           & \zqh{85.17}   & \zqh{68.08}      \\
		& \zqh{Sentic GCN \cite{liang2022aspect}}    & \zqh{77.90}       & \zqh{74.71}      & \zqh{84.03}     & \zqh{75.38}     & \zqh{-}           & \zqh{-}        & \zqh{\underline{82.84}}           & \zqh{\underline{67.32}}          & \zqh{\textbf{90.88}}   & \zqh{\underline{75.91}}     \\
		& \textbf{KGAN (Ours)}    & \textbf{78.91}     & \textbf{75.21}   & \textbf{84.46}    & \underline{77.47} & \zqh{\textbf{78.55}}      & \zqh{\textbf{77.45}}       & \zqh{\textbf{83.09}}           & \zqh{\textbf{67.90}}          & \zqh{89.78}   & \zqh{74.58}      \\ \midrule
		\multirow{8}{*}{BERT}      
        & \zqh{Vanilla BERT$^\ddag$ \cite{devlin2018bert}} & \zqh{77.58}     & \zqh{72.38}    & \zqh{85.62}       & \zqh{78.28}       & \zqh{75.28}   & \zqh{74.11}     & \zqh{83.40}    & \zqh{65.28}    & \zqh{89.54}   & \zqh{70.47}     \\
		& R-GAT-BERT \cite{wang2020relational}  & 78.21    & 74.07     & 86.60     & 81.35     & 76.15   & 74.88      & \zqh{83.22}           & \zqh{69.73}          & \zqh{89.71}   & \zqh{76.62}   \\
		& DGEDT-BERT \cite{tang2020dependency}    & 79.80    & 75.60    & 86.30     & 80.00       & 77.90     & 75.40     & \zqh{84.00}           & \zqh{71.00}          & \zqh{91.90}   & \zqh{79.00}      \\
		& \zqh{T-GCN-BERT~\cite{tian2021aspect}}     & \zqh{80.88}     & \zqh{77.03}     & \zqh{86.16}         & \zqh{79.95}     & \zqh{76.45}    & \zqh{75.25}     & \zqh{85.26}           & \zqh{\underline{71.69}}          & \zqh{\underline{92.32}}   & \zqh{77.29}     \\
		& DM-GCN-BERT \cite{pang-etal-2021-dynamic}    & 80.22  & 77.28     & \textbf{87.66}    & \textbf{82.79}    & \underline{78.06}    & \underline{77.36}      & \zqh{-}      & \zqh{-}   & \zqh{-}       & \zqh{-}        \\
		& DualGCN-BERT \cite{li-etal-2021-dual-graph}     & \underline{81.80}     & \underline{78.10}    & 87.13   & 81.16   & 77.40      & 76.02      & \zqh{-}      & \zqh{-}   & \zqh{-}       & \zqh{-}                     \\
		& \zqh{Sentic GCN-BERT \cite{liang2022aspect}}     & \zqh{82.12}    & \zqh{79.05}    & \zqh{86.92}     & \zqh{81.03}      & \zqh{-}     & \zqh{-}     & \zqh{\underline{85.32}}           & \zqh{71.28}          & \zqh{91.97}   & \zqh{\underline{79.56}}      \\
		& \textbf{KGAN-BERT (Ours)}   & \textbf{82.66}    & \textbf{78.98}     & \underline{87.15} & \underline{82.05} & \zqh{\textbf{79.97}}   & \zqh{\textbf{79.39}}    & \zqh{\textbf{86.21}}   & \zqh{\textbf{74.20}}          & \zqh{\textbf{92.34}}   & \zqh{\textbf{81.31}}     \\ \midrule
		\multirow{4}{*}{RoBERTa}                                                                                 
        & \zqh{Vanilla RoBERTa$^\S$ \cite{liu2019roberta}} & \zqh{\underline{83.78}} & \zqh{\underline{80.73}} & \zqh{87.37}       & \zqh{80.96}       & \zqh{\underline{77.17}} & \zqh{\underline{76.20}}  & \zqh{\underline{84.56}}   & \zqh{\underline{70.16}} & \zqh{\underline{91.25}}    & \zqh{\underline{73.44}}    \\
		& \zqh{ASGCN-RoBERTa$^\S$ \cite{zhang2019aspect}}       & \zqh{83.33}   & \zqh{80.32}  & \zqh{86.87}       & \zqh{80.59}       & \zqh{76.10}        & \zqh{75.07}          & \zqh{-}      & \zqh{-}   & \zqh{-}       & \zqh{-}    \\
		& RGAT-RoBERTa$^\S$ \cite{wang2020relational}     & 83.33     & 79.95      & \underline{87.52} & \underline{81.29} & 75.81   & 74.91    & \zqh{-}      & \zqh{-}   & \zqh{-}       & \zqh{-}      \\
		& \textbf{KGAN-RoBERTa (Ours)}      & \textbf{83.91}     & \textbf{81.07}    & \textbf{88.45}    & \textbf{84.05}    & \zqh{\textbf{80.55}}   & \zqh{\textbf{79.63}}      & \zqh{\textbf{88.60}}      & \zqh{\textbf{74.36}}   & \zqh{\textbf{94.38}}       & \zqh{\textbf{83.23}}      \\
        \Xhline{1.2pt}
	\end{tabular}
\end{table*}